\documentclass[11pt,a4paper]{article}
\usepackage[final]{acl2022}
\usepackage{times}
\usepackage[flushleft]{threeparttable}
\usepackage{todonotes}
\usepackage{enumitem}
\usepackage{algorithm}
\usepackage{algpseudocode}
\usepackage{booktabs}
\usepackage{subcaption}
\usepackage[nointegrals]{wasysym}
\usepackage{tikz}
\usepackage{tabu}
\usepackage{tabularx}
\usepackage{float}
\usepackage{listings}
\usepackage{color}
\usepackage{multirow}
\usepackage{colortbl}
\usepackage{framed}
\usepackage{pifont}
\usepackage{dsfont}
\usepackage{amsmath,amsthm}
\usepackage{amssymb}
\usepackage{graphicx}
\usepackage{textcomp}
\usepackage{tcolorbox}
\usepackage[table]{xcolor}
\usepackage{soul}
\usepackage{lipsum}
\usepackage{tabularx}
\usepackage{microtype}

\usepackage{titlesec}
\titlespacing*{\section}{0pt}{*0.8}{*0.8}
\titlespacing*{\subsection}{0pt}{*0.8}{*0.8}

\definecolor{light-gray}{gray}{0.9}

\newcommand{\temprel}[1]{$\langle$\textit{#1}$\rangle$}

\title{Go Back in Time: \\Generating Flashbacks in Stories with Event Temporal Prompts}

\author{Rujun Han $^{1}$ ~ Hong Chen$^{2}$ ~ Yufei Tian$^{3}$ ~ Nanyun Peng$^{3}$ \\
$^1$University of Southern California ~ $^2$Tokyo University \\ $^3$University of California, Los Angeles \\
{\tt rujunhan@usc.edu;  chen@nlab.ci.i.u-tokyo.ac.jp} \\
{\tt yufeit@ucla.edu;  violetpeng@cs.ucla.edu}
}

\date{}

\begin{document}
\maketitle

\begin{abstract}
Stories or narratives are comprised of a sequence of events.
To compose interesting stories, professional writers often leverage a creative writing technique called \textit{flashback} that inserts past events into current storylines as we commonly observe in novels and plays. However, it is challenging for machines to generate \textit{flashbacks} as it requires solid understanding of event \textbf{temporal order} (e.g. \textit{feeling hungry} \temprel{before} \textit{eat}, not vice versa), and the creativity to arrange storylines so that earlier events do not always appear first in \textbf{narrative order}. 
Two major issues in existing systems exacerbate the challenges: 1) temporal bias in pretraining and story datasets that leads to monotonic event temporal orders; 2) lack of explicit guidance that helps machines decide where to insert \textit{flashbacks}. We propose to address these issues using structured storylines to encode events and their pair-wise temporal relations (\temprel{before}, \temprel{after} and \temprel{vague}) as \textbf{temporal prompts} that guide how stories should unfold temporally. We leverage a Plan-and-Write framework enhanced by reinforcement learning to generate storylines and stories end-to-end. Evaluation results show that the proposed method can generate more interesting stories with \textit{flashbacks} while maintaining textual diversity, fluency and temporal coherence.\footnote{Code, data and trained models are available here: \url{https://github.com/PlusLabNLP/flashback\_gen}}
\end{abstract}

\section{Introduction}
\begin{figure}[t]
    \begin{subfigure}[b]{\columnwidth}
    \centering
    \includegraphics[trim=0cm 0cm 0cm 0cm, clip, width=\columnwidth]{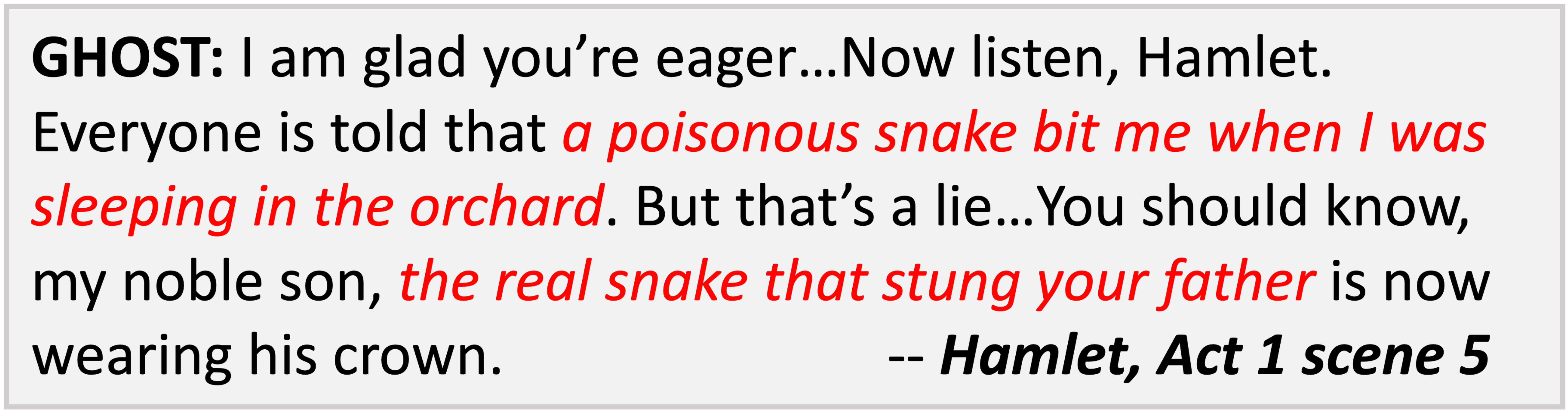}
    \caption{A \textit{flashback} example from William Shakespeare's famous play \textit{Hamlet} (in plain English). Red text indicates past events.}
    \label{fig:illustrating-exp}
    \vspace{0.2cm}
    \end{subfigure}
    ~
    \begin{subfigure}[b]{\columnwidth}
    \includegraphics[trim=0cm 0cm 0cm 0cm, clip, width=\columnwidth]{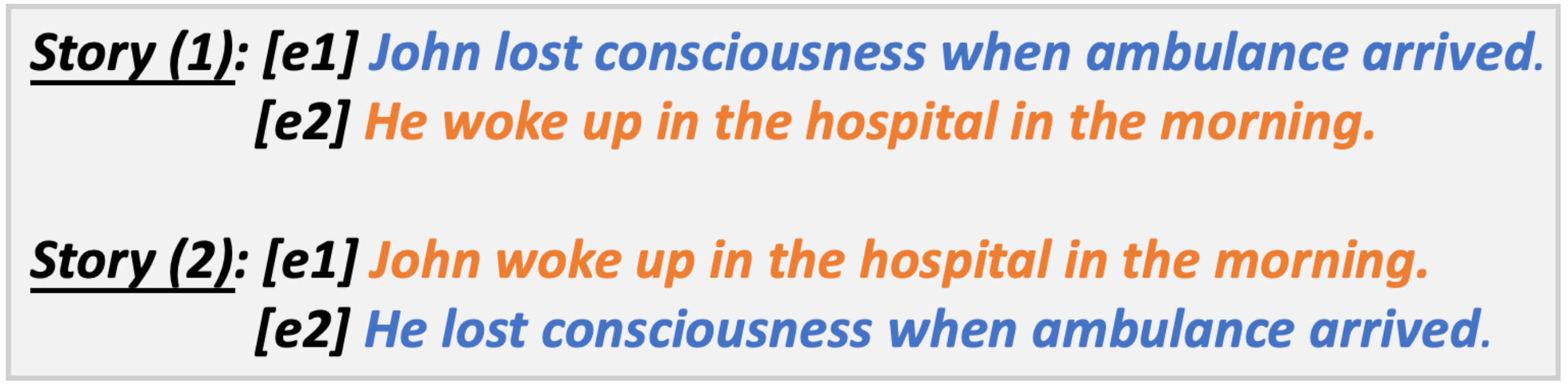}
    \caption{Two-sentence stories with the same event \textbf{temporal order} but different \textbf{narrative order}. The second one with a flashback is intuitively more interesting than the first one.}
    \label{fig:illustrating-order}
    \end{subfigure}
 \caption{(a) \textit{flashback} (b) temporal v.s. narrative order.}  
 \vspace{-0.7cm}
\label{fig:examples}
\end{figure}

\textit{Flashback} is a popular creative writing technique that brings the readers from the present moment to the past via inserting earlier events in order to provide background or context of the current narrative \citep{flashback-pavis, flashback-kenny, flashback-gebeyehu}. For example, in Figure~\ref{fig:illustrating-exp}, the ``GHOST'' in Shakespeare's play \textit{Hamlet} instruments a \textit{flashback} by interrupting the main narrative and describing a historical event to the audience that \textit{Hamlet}'s father was killed by the current king rather than a snake.

\textit{Flashback}, by manipulating the event temporal orders in narrative structure, can arouse readers' emotions such as surprise, suspense, and curiosity \citep{flashback-Event-Schemas, flashback-Stories}. These emotions stimulate readers' interests and eventually contribute to the satisfaction of reading \citep{flashback-Emotion, flashback-Maintaining}, which improves the \textbf{interest level} of a story. The example in Figure~\ref{fig:illustrating-exp} injects historical events in the middle of the narrative. This arrangement of events can surprise readers and therefore, makes the story more interesting than a straightforward storyline where the past events are shown in the beginning.

Similarly, consider the pair of two-sentence stories in Figure~\ref{fig:illustrating-order}. Both stories are composed of the same events with the \textbf{temporal order} \textit{``lost consciousness''} \temprel{before} \textit{``woke up in the hospital.''} In Story (1), seeing \textbf{[e1]}, readers can make a relatively easy educated guess of \textbf{[e2]}, but it is more subtle in Story (2) as there are many different ways to end up in a hospital. By showing the ending event first, the \textit{flashback} in Story (2) creates suspense that makes the following sentences less predictable, and thus arouses readers' curiosity and makes the reading more interesting.

While human writers are capable of maneuvering event temporal orders to compose coherent and interesting stories, it remains challenging for machines. The challenge is partially attributed to data bias. \citet{ning-etal-2018-improving} shows that the pattern in Story (1) is dominant in human-written texts, where neighboring events with \temprel{before} temporal relations (i.e., \textbf{narrative order} indicates \textbf{temporal order}) occur $60-70\%$ of the time. This is also manifested in our experiments with vanilla language models amplifying this ratio and producing more than 80\% \temprel{before} relations for neighboring events in the generated stories. 
Furthermore, current state-of-the-art story generation systems that incorporate event representations usually assume event \textbf{temporal order} follows \textbf{narrative order} \citep{goldfarb-tarrant-etal-2020-content, lin-etal-2021-conditional}. There are no explicit prompts in these systems that help determine when \textit{flashback} should be used, leaving models to produce dull stories consisting of event sequences with monotonic \temprel{before} relations. 

\begin{figure*}[t]
    \centering
\includegraphics[trim={6cm 0.1cm 6cm 0.0cm}, clip, angle=-90, width=0.9\textwidth]{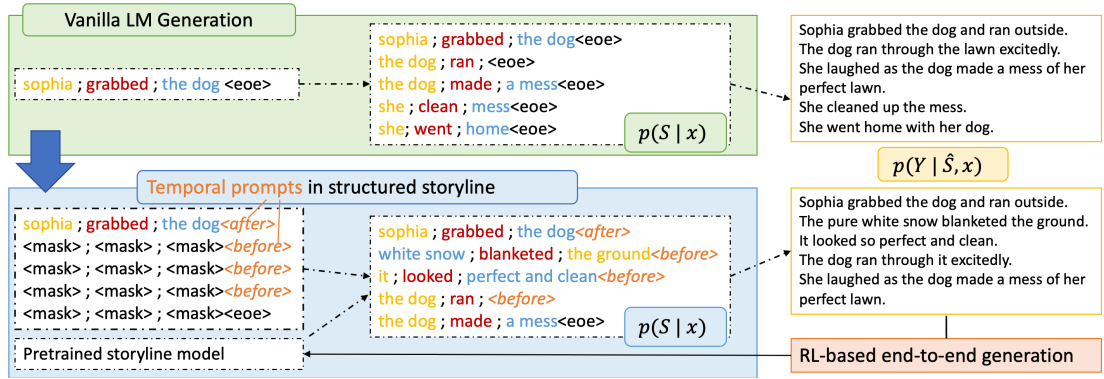}
\caption{An illustration of our overall model. Here we use the first sentence of the story (and its associated event representation) as input $\boldsymbol{x}$. The upper block shows the vanilla implementation of the Plan-and-Write workflow. The bottom block is our core novel design by leveraging \textbf{temporal prompts} in structured storylines to generate \textit{flashbacks}. For illustration purposes, we re-order the triggers and arguments, and storylines are ground-truths (i.e. not predicted by models). Our final model uses reinforcement learning to implement end-to-end training.}
\label{fig:model}
\vspace{-0.5cm}
\end{figure*}

To facilitate more effective \textit{flashback}, we propose to incorporate \textbf{temporal prompts} in an end-to-end story generation framework inspired by the Plan-and-Write paradigm~\citep{yao2019plan, xu-etal-2020-megatron, goldfarb-tarrant-etal-2020-content}, where machines first learn to plot a storyline, and then generate the story based on the storyline. Specifically, we encode predefined event \textbf{temporal prompts} in structured storylines. As the bottom block of Figure~\ref{fig:model} shows, a structured storyline contains two components: 1) \textbf{event representations} where an event trigger (``grabbed'') and two arguments (``she'' and ``the dog'') are extracted from the original story sentences; and 2) \textbf{temporal prompts}: the temporal order between neighboring events, e.g. event 1: (``she'', ``grabbed'', ``the dog'') is \temprel{after} event 2: (``white snow'', ``blanketed'', ``the ground''). By training our storyline generation model with these predefined pair-wise temporal relations, models capture how neighboring events are temporally related to each other; while during storyline decoding, supplying predefined \textbf{temporal prompts} can guide models to generate reasonable narratives with desirable event temporal orders.

Prior works~\citep{fan-etal-2019-strategies,goldfarb-tarrant-etal-2020-content} build the storyline and story models separately, which creates a discrepancy where gold storylines are used during training, but predicted storylines are used during inference.
To mitigate this training-inference discrepancy, we leverage reinforcement learning (RL) to train our systems end-to-end. It enables the story model to train on generated storylines and updates the storyline model with the feedback from the story model. Our experimental results show that the RL-based models can leverage \textbf{temporal prompts} more effectively, resulting in more effective \textit{flashback} generation and more interesting stories.

We summarize the contributions of this paper as follows: 1) To facilitate effective \textit{flashback}, we propose to leverage structured storylines with \textbf{temporal prompts} to arrange events in story generation. 2) We integrate reinforcement learning in our story generation pipeline, which can help models better leverage \textbf{temporal prompts}. 3) We test our framework on two open-domain story datasets and show more effective \textit{flashbacks} and increased \textbf{interest level} while maintaining fluency and temporal commonsense in the generated stories. To our best knowledge, this is a pioneering study on \textit{flashbacks} in \textit{neural story generation}.

\section{Background and Task Definitions}
\label{sec:task-define}
In this section, we describe the key components: \textbf{events} and \textbf{temporal prompts} in our proposed structured storylines, and then define the Plan-and-Write generation task.
\vspace{-0.1cm}
\paragraph{Event Representation.} Following the definitions of \citet{ACE}, we define an event as a trigger word and its arguments. In this work, we simplify the representation by leveraging semantic role labeling (SRL) tools \citep{Shi2019SimpleBM} to parse two arguments as shown in Figure~\ref{fig:model}. We only consider one event per story sentence and denote the $k$-th event in story $i$ as $e_{i, k}$. We leave more complicated representations for future study.
\vspace{-0.1cm}
\paragraph{Temporal Prompts.} Let $\boldsymbol{r}_i = \{r_{i, k}\}$ denotes the set of temporal relations between the $k$-th and the $(k+1)$-th event in story $i$.  
If $k$ indexes the last event, $r_{i, k}$ is not defined. Following the event relation definition of \cite{ning-etal-2018-multi}, we use events' \textit{start time} to evaluate temporal order. 
\vspace{-0.1cm}
\paragraph{Structured Storyline.} Figure~\ref{fig:model} provides a storyline consisting of five event representations extracted from our data. More formally, let $\mathcal{S}_i = \{e_{i,1}, e_{i,2}, ... e_{i, k}, ... e_{i, n}\}$ indicates a storyline with $n$ events. Encoding \textbf{temporal prompts}, $\mathcal{S}_i$ becomes $\mathcal{S}^r_i = \{e_{i,1}, r_{i,1}, e_{i,2}, r_{i,2} ... e_{i, k}, r_{i, k}, ... e_{i, n}\}$. Note that in this work, $\boldsymbol{r}_i$ \textbf{is provided as predefined prompts} rather than predicted as $e_{i, k}$.
\vspace{-0.1cm}
\paragraph{Story.} Our ultimate goal is to generate \textit{flashbacks} in stories. We denote the story associated with the storyline $\mathcal{S}_i$ as $\mathcal{Y}_i$. 

\paragraph{Plan-and-Write} is a two-stage framework that first generates storyline $\hat{\mathcal{S}_i}$ given some input $\boldsymbol{x}$ (e.g. title, leading sentence), and then generate $\hat{\mathcal{Y}_i}$ based on $\hat{\mathcal{S}_i}$. Again, $r_{i, k}$ are given as \textit{predefined} prompts whereas $e_{i, k}$ are to be predicted as part of the storyline generation shown in Figure~\ref{fig:model}.
\section{Framework for FlashBack Generation} 
\label{sec:method}

In this section, we first provide an overview of the Plan-and-Write story generation system and introduce a vanilla version of the end-to-end training method. Then we describe the details of our key contribution of leveraging event \textbf{temporal prompts} to generate \textit{flashbacks}. After that, we discuss pretraining structured storylines with self-labeled data and incorporating reinforcement learning to jointly train our end-to-end models.

\subsection{Plan-and-Write Models}
\label{sec:base-model}

In order to provide better explainability and controllability over the machine generated stories, recent research efforts \citep{yao2019plan, xu-etal-2020-megatron, goldfarb-tarrant-etal-2020-content} explore dividing story generation into two steps: 1) from input or prefix, $\boldsymbol{x}$, we first produce a storyline, $\mathcal{S}_i$; 2) based on the storyline, we generate a story, $\mathcal{Y}_i$. We describe the details below.
\vspace{-0.1cm}
\paragraph{Storyline Model.} Let $\alpha$ denote the parameters of the storyline model, per sample training loss can be computed as
$\mathcal{L}_{\alpha} = -\log p\left(\mathcal{S}_i|\boldsymbol{x}_i, \alpha \right)$.
\vspace{-0.1cm}
\paragraph{Story Model.} Let $\beta$ denote the parameters of the story model, per sample training loss can be computed as $
\mathcal{L}_{\beta} = -\log p\left(\mathcal{Y}_i|\boldsymbol{x}_i, \mathcal{S}_i, \beta \right)$.
\vspace{-0.1cm}
\paragraph{Inference.} Note that $\mathcal{S}_i$ above is the gold storyline extracted from $\mathcal{Y}_i$. At the inference time, we do not have $\mathcal{S}_i$, and have to replace it with $\hat{\mathcal{S}_i}$, the predicted storyline. This results in a discrepancy between the training and inference time.
\vspace{-0.1cm}
\paragraph{End-to-end Training.} 
Instead of using gold storyline $\mathcal{S}_i$ to train a story model, we can take $\hat{\mathcal{S}_i}$ as its input. Now the per sample training loss for the story model becomes $
\mathcal{L}_{\theta} = -\log p\left(\mathcal{Y}_i|\boldsymbol{x}_i, \hat{\mathcal{S}_i}, \theta \right)$,
where $\theta$ indicates the end-to-end story model parameters. End-to-end training can alleviate the gap between the training and inference time, and potentially lead to more consistent stories \footnote{End-to-end training may have small coverage of the generated events in stories, which we address in Appendix~\ref{sec:tradeoff}.}.

\subsection{Structured Storyline Construction}
\label{sec:struct-input}
As Figure~\ref{fig:model} shows, for a story sentence, we first use the SRL tool to parse its trigger ${t_{i,k}}$ and two arguments ${a^1_{i,k}}$ and ${a^2_{i,k}}$. We then convert this representation into a textual form: ``${t_{i,k}}$ ; ${a^1_{i,k}}$ ; ${a^2_{i,k}}\langle eoe \rangle$'', where ``;'' separates two event components, and $\langle eoe \rangle$ indicates event ending. For example, the parsed ${t_{i,k}}$, ${a^1_{i,k}}$ and ${a^2_{i,k}}$ in the story sentence ``she grabbed the dog and ran outside'' are ``grabbed,'' ``she'' and ``the dog'' respectively. They are concatenated into a final textual representation as ``grabbed ; she ; the dog$\langle eoe \rangle$.''

Depending on the experimental setup, we may use no or only the leading event as input, $\boldsymbol{x}$. Inspired by the mask prediction design in \citet{devlin-etal-2019-bert, ROBERTA-19, lewis-etal-2020-bart}, we represent the remaining missing events in the inputs as ``$\langle$mask$\rangle$ ; $\langle$mask$\rangle$ ; $\langle$mask$\rangle$ ; $\langle eoe \rangle$,'' where $\langle$mask$\rangle$ indicates either event trigger word or arguments to be predicted by the storyline model.

\subsection{Temporal Prompt Encoding}
\label{sec:temp-prompt}
Temporal prompts are used to generate \textit{flashbacks}.  As we mentioned in Section~\ref{sec:task-define}, we encode a sequence of \textit{predefined} event \textbf{temporal prompts} $\boldsymbol{r}_i = \{r_{i,k}\}$ in storyline for $k \in \{1, n-1\}$ to help models determine whether the next event mention (in narrative order) should start earlier or later than its preceding event mention. We use temporal relation extraction tools to annotate all $r_{i, k}$ in our experimental data. 
Specifically, we use ECONET \citep{han-etal-2021-econet} finetuned on the MATRES dataset \citep{ning-etal-2018-multi} to predict the temporal relation between neighboring events.\footnote{The ECONET tool is available here: \url{https://github.com/PlusLabNLP/ECONET}.} The context and the locations of a pair of event trigger words are fed into ECONET to predict their temporal order.
The \textbf{temporal prompt} set consists of \temprel{before}, \temprel{after} and \temprel{vague} (capturing undetermined temporal order), and are fixed in $S_i^r$. 
Note that \temprel{vague} indicates undetermined temporal order due to the ambiguity of the context \cite{cassidy-etal-2014-annotation, ning-etal-2018-multi} and it does not suggest the context is poor or the relations are wrong. As shown in Figure~\ref{fig:model}, we replace the end-of-event token $\langle eoe \rangle$ with \textbf{temporal prompts} in storylines, except for the last event which does not have a next event. With the prompt-augmented storylines, $S_i^r$, we can re-write the storyline loss as $\mathcal{L}_{\alpha} = -\log p\left(\mathcal{S}^r_i|\boldsymbol{x}_i, \boldsymbol{r}_i, \alpha \right)$, and story loss as $\mathcal{L}_{\theta} = -\log p\left(\mathcal{Y}_i|\boldsymbol{x}_i, \hat{\mathcal{S}^r_i}, \theta \right)$.

\begin{algorithm}[t]
\small
\caption{RL-based End-to-end Training}
\begin{algorithmic}[1]
\State Randomly initialize $\alpha$ and $\theta$
\State Pretrain $\alpha$  \Comment{storyline pretraining}
\For{$i \in \mathbf{M}$} \Comment{loop through all data\footnote{actual implementation uses batch updates}}
    \State Generate $\hat{S_i^r}$ from storyline model ($\alpha$)
    \State $\mathcal{L}_{\theta} = -\log p\left(\mathcal{Y}_i|\boldsymbol{x}_i, \hat{\mathcal{S}^r_i}, \theta \right)$
    \State $\nabla J_\alpha = R_i \cdot \nabla \log \left(p(S_i | \boldsymbol{x}_i, \boldsymbol{r}_i, \alpha) \right)$
    \State $\alpha = \alpha - \nabla J_\alpha$
    \State $\theta = \theta - \nabla \mathcal{L}_{\theta}$
\EndFor
\end{algorithmic}
\label{algo:1}
\end{algorithm}

\subsection{Storyline Pretraining}
\label{sec:pretrain}
Using intermediate pretraining to adapt original pretrained language models has been shown to be effective for a variety of downstream tasks such as information extraction \cite{joshi-etal-2020-spanbert}, question-answering \citep{khashabi-etal-2020-unifiedqa, Garg_2020} and commonsense reasoning \citep{zhou2020pre}. To capture more diverse event sequences and facilitate better story generation, we explore pretraining storyline model with SRL extracted storyline from BookCorpus dataset \citep{Zhu_2015_ICCV}, and use learned $\alpha$ to initialize storyline models.

\subsection{RL-based End-to-end Model}
\label{sec:rl-model}
The end-to-end model described in Sec.~\ref{sec:base-model} allows the story model to train with the generated storylines and hence alleviate the gap between training and inference. However, this workflow still lacks a mechanism that enables the storyline model to adjust with the feedback from the story model. The challenges of training storyline and story models jointly originate from decoding storylines as inputs for the story model, which involves non-differentiable token selections. Thus, the final loss $L_\theta$ cannot be directly back-propagated into the storyline model.  

To overcome this barrier, we adopt reinforcement learning (RL), specifically, the REINFORCE algorithm \citep{Williams91functionoptimization} in our end-to-end training. Let $R_i = R(\boldsymbol{x}_i, \boldsymbol{r}_i)$. The expected reward with respect to the storyline model can be written as $\mathbb{E}_\alpha\left[R_i \right] = \mathbb{E}\left[R_i \cdot \log \left(p(S^r_i | \boldsymbol{x}_i, \boldsymbol{r}_i, \alpha) \right)\right]$. The gradient to update the storyline model is $\nabla J_\alpha = \mathbb{E}\left[R_i \cdot \nabla \log \left(p(S^r_i | \boldsymbol{x}_i, \boldsymbol{r}_i, \alpha) \right)\right]$, which can be approximated with sampling techniques. Motivated by \citet{xu-etal-2018-skeleton}, we use negative loss of the story model to construct rewards, that is, $R = - \mathcal{L}_\theta$.\footnote{We do not use baseline reward as we found this simple reward design works effectively in our experiments.} In other words, smaller loss from the story model is associated with larger reward. Algorithm~\ref{algo:1} summarizes the overall method.
\section{Experimental Setup}
\label{sec:implement}
In this section, we start by describing our research objectives, then we describe our data, evaluation metrics, experimental designs and implementation details aiming to achieve these objectives.

\paragraph{The overall research objective} is to measure the impact of using \textbf{temporal prompts} in structured storylines. Specifically, can \temprel{after} successfully induce \textit{flashbacks}? If so, does that contribute to the interest level of the generated stories while maintaining the overall quality of the texts? 

\subsection{Datasets.} ROCStories \citep{mostafazadeh-etal-2016-corpus} and WritingPrompts \cite{fan-etal-2018-hierarchical}  are our experimental datasets. We ensured all reported results using the same test data as the baseline systems \citep{xu-etal-2020-megatron} and \cite{goldfarb-tarrant-etal-2020-content}. For pretraining data, we use BookCorpus \citep{Zhu_2015_ICCV}. Appendix~\ref{sec:append-data} shows all details of data splits and pre-processing process.

\subsection{Temporal Prompts Constructions}
ECONET was finetuned three times with different random seeds, so we take the consensus vote from three models. If there is any disagreement, we label the temporal order as \temprel{vague}. We benchmark ECONET's annotation performances in Appendix~\ref{sec:benchmark}, which shows it provides highly accurate temporal relations.
For human evaluations specifically, we consider two prompt settings in order to gauge different impacts of \temprel{after}. 
1) for ROCStories, all structured storylines consist of exactly four \textit{predefined} \textbf{temporal prompts} created following Sec~\ref{sec:temp-prompt}. We randomly sample stories with one \temprel{after} prompt from the test data. We will show later in the analysis that vanilla language models would generate more than 80\% event pairs with \temprel{before} relations for ROCStories; \temprel{after} prompt should bring this ratio down if it is effective. 2) for WritingPrompts, since the number of events is not fixed, we randomly sample test stories generated with \temprel{after} prompts for evaluation.

\subsection{Automatic Evaluation Metrics}
We use automatic metrics to evaluate the textual quality of stories. We report \textbf{Ref. PPL}: reference stories' perplexity in our models and \textbf{Gen. PPL}: generated stories' perplexity scored by GPT-2 \citep{radford2019language}. For diversity, we report \textbf{Distinct Ratio} (\%): overall vocabulary:token number ratio. We also report standard \textbf{BLEU-3} and \textbf{ROUGE}$_L$.

\subsection{Human Evaluation Metrics} 
We rely on human annotators to analyze the effectiveness of \textit{flashback} generations. We request 18 MTurkers who succeeded in our previous annotation tasks \citep{han-etal-2021-ester} to evaluate stories produced by our compared models. We host a small qualification round followed by a large annotation task. Only 10 workers are qualified, and we only consider their annotations. Eventually, we collect 106 and 77 sets of valid annotations for ROCStories and WritingPrompts.
\vspace{-0.2cm}
\paragraph{Temporal diversity.} The dominance of \temprel{before} relation in our data can make models biased toward generating stories with more \temprel{before} relations. Therefore, we are interested to see how inserting an \temprel{after} prompt can help \textbf{increase} the percentage of non-\temprel{before} event relations in the generated stories. Let $\hat{R}_{r}$ indicate the percentage of a particular relation annotated by MTurkers. We calculate the entropy of the set $\{\hat{R}_{r}\}, \forall r \in$ $\{$\temprel{before}, \temprel{after}, \temprel{vague}$\}$ to measure temporal diversity.
\vspace{-0.2cm}
\paragraph{Accuracy} measures the percentage of \temprel{after} being correctly incorporated in the generated stories labeled by human annotators. We used a relaxed version by counting annotated \temprel{vague} as correct too, as \temprel{vague} can potentially be \temprel{after}. Both \textbf{accuracy} and \textbf{temporal diversity} can show the effectiveness of generating \textit{flashbacks} using \temprel{after}.
\vspace{-0.5cm}
\paragraph{Temporal coherence} indicates if the event sequence in a generated story aligns with an annotator's \textbf{temporal commonsense}.\footnote{interface with detailed instructions and examples can be found in Figure~\ref{fig:instruct_2} of the appendix.} 1 and 0 correspond to yes and no, respectively.
\vspace{-0.2cm}
\paragraph{Interest level.} Precisely defining interest level is difficult as it is a broad concept. So we focus on the \textit{unexpectedness} component of cognitive interest. As pointed out by \citet{behrooz-thesis}, \textit{unexpectedness} can be further explained as how predictive an event is, which is closely related to \textit{flashback} generation. Therefore, we define an \textit{interesting} event as 1) being unexpected or surprising and 2) being logical according to the context and general commonsense.\footnote{the second definition is not identical to the temporal coherence above. Events contradicting general commonsense can still be temporally coherent (see Figure~\ref{fig:instruct_2}) for examples.}

For the compared models, we ask annotators to provide ranks between 1 to K for the generated stories, with K indicating the most interesting story and 1 indicating the least interesting one. We encourage workers to provide different scores for all compared stories, but equal scores are allowed. The max score K depends on the number of compared models, 5 for ROCStories and 4 for WritingPrompts. We provide detailed instructions in the interface shown in the appendix. Crucially, \textbf{interest level} is separately annotated from other metrics and we ensure annotators do not see the same set of stories in both tasks.

\begin{table*}[htbp!]
\centering
\small
\scalebox{0.78}{
\setlength{\tabcolsep}{4pt}
\begin{tabular}{l|ccccc|cccc}
\toprule
 & \multicolumn{5}{c|}{\textbf{Automatic Evaluation}} & \multicolumn{4}{c}{\textbf{Human Evaluation}} \\
 \midrule
 & \textbf{Ref.} & \textbf{Gen.} & \textbf{Distinct} & \textbf{BLEU} & \textbf{ROUGE}$_L$ & \textbf{Temporal} & \textbf{Accuracy} & \textbf{Temporal} & \textbf{Interest} \\
\textbf{Models} & \textbf{PPL}($\downarrow$) & \textbf{PPL}($\downarrow$) & \textbf{Ratio}($\uparrow$) &  ($\uparrow$) &  ($\uparrow$)  & \textbf{Diversity}($\uparrow$) & ($\uparrow$) & \textbf{Coherence} ($\uparrow$) & \textbf{Level} ($\uparrow$)\\
\midrule
\textsc{Temporal-BART} & 24.65  & 19.47 & 4.10 & 5.01 & 19.12 & - & - & - & -\\
\textsc{MEGATRON} & - & 34.14 & \textbf{4.57} & 2.57 & 15.23 & \textbf{1.21}* & - & 0.78 & 2.69\\
\midrule
\textsc{Vanilla-Gen} & 27.30 & 19.29 & 3.99 & 5.13 & 19.29 & 0.88 & - & \textbf{0.88} & 2.95 \\
\textsc{+ Structured-Prompt} & 22.85 & 19.94 & 4.09 & 5.07 & 19.39 & 1.09 & 55.75 & 0.82 & 3.03 \\
\textsc{+ Pretrained} & 21.16 & \textbf{19.25} & 4.01 & 5.06 & 19.44 & 1.07 & 52.21 & 0.84 & 2.96\\
\textsc{+ RL} & \textbf{15.45} & 19.42 & 4.17 & \textbf{5.20} & \textbf{19.49} & 1.14 & \textbf{56.64} & 0.86 & \textbf{3.06}\\
\bottomrule
\end{tabular}
}
\vspace{-0.3cm}
\caption{Evaluation results for ROCStories. All values in the bottom block are averaged over three runs. \textsc{MEGATRON}'s \textbf{Ref. PPL} is missing as it is not reported in the paper and the model is not publicly available. \textsc{MEGATRON} and \textsc{Vanilla-Gen} do not use any \textbf{temporal prompts}, so \textbf{accuracy} is not applied. \textsc{MEGATRON}'s \textbf{temporal diversity}* is high due to large amount of \temprel{vague} annotations in its generated stories.}
\label{tab:res-roc}
\vspace{-0.5cm}
\end{table*}

\begin{table}[t]
    \small
    \setlength{\tabcolsep}{1pt}
    \scalebox{0.8}{
 	\centering
 	\setlength\extrarowheight{1pt}
 	\begin{tabular}{ll} \hline
 	\toprule
 	\textbf{Input} & [s1] my favorite class was english literature. \\
 	\textbf{prompt} & \temprel{before} \textbf{\temprel{after}} \temprel{before} \temprel{before}  \\
 	\midrule
 	\rowcolor{light-gray}  & [s2] i was in english class for the first time. \\
 	\rowcolor{light-gray} & [s3]  i had to take a class with my local author this semester.   \\ 
 	\rowcolor{light-gray} \multirow{-2}{*}{\textbf{baseline}} & [s4] i decided to ask him to speak to me at a local book festival.  \\
 	\rowcolor{light-gray} & [s5] he said that he was really good and i liked him. \\ 
      & [s2] one day, i decided to go to the library. \\
 	\multirow{2}{*}{\textsc{Vanilla}} & [s3] i found a book that i really liked. \\
      & [s4] i read the whole book in one day. \\
     & [s5] i was so happy that i went back to school.\\
 	\rowcolor{light-gray}  & [s2] i would go to the library to read all the books. \\
 	\rowcolor{light-gray} & [s3] \hl{i had never been to a library before.}   \\ 
 	\rowcolor{light-gray} \multirow{-2}{*}{\textsc{RL}} & [s4] i was so excited to go.  \\
 	\rowcolor{light-gray} & [s5] i ended up spending a lot of time in the library.  \\ 
 	
    \toprule
 	\textbf{Input} & [s1] tom was excited to go on a date. \\
 	\textbf{prompt} & \textbf{\temprel{after}} \temprel{before}  \temprel{before} \temprel{before}  \\
 	\midrule
 	\rowcolor{light-gray}  & [s2] he went to his favorite restaurant to pick up his date. \\
 	\rowcolor{light-gray} & [s3] the date went well and then tom got drunk.   \\ 
 	\rowcolor{light-gray} \multirow{-2}{*}{\textbf{baseline}} & [s4] he told his date they were going to a comedy show.  \\
 	\rowcolor{light-gray} & [s5] the date went on and they had a great time at the movie. \\ 
      & [s2] he was going to take his girlfriend to the movies. \\
 	\multirow{2}{*}{\textsc{Vanilla}} & [s3] they went to the movie theater. \\
      & [s4] they had a great time. \\
     & [s5] tom couldn't wait to go back again.\\
 	\rowcolor{light-gray}  & [s2] \hl{he had been looking forward to it for months.} \\
 	\rowcolor{light-gray} & [s3] he went to the mall and bought a new dress.   \\ 
 	\rowcolor{light-gray} \multirow{-2}{*}{\textsc{RL}} & [s4] he wore it to his date's house.  \\
 	\rowcolor{light-gray} & [s5] his date loved it and they had a great time. \\ 
 	\bottomrule
 	\end{tabular}
 	}
 	\vspace{-0.2cm}
   	\caption{Generated stories by our compared models. baseline: \textsc{MEGATRON}, \textsc{Vanilla-Gen} and \textsc{RL}-based. end-to-end. Input is the first sentence. Highlighted sentences contain an event that started earlier than the preceding event.}
   	\label{tab:roc-examples}
   \vspace{-0.7cm}
 \end{table}

\subsection{Compared Models} 

\paragraph{Baselines.} \citet{xu-etal-2020-megatron}, denoted as \textsc{MEGATRON}, is chosen as the baseline as it outperforms previous systems such as \citet{guan-etal-2020-knowledge} on ROCStories. We also compare with \textsc{TemporalBART} \citep{lin-etal-2021-conditional} as it is pretrained with temporal ordering and event infilling tasks. For WritingPrompts, we compare with \textsc{ContentPlanning} \citep{goldfarb-tarrant-etal-2020-content} as it also adopts the Plan-and-Write workflow as well as structured event representations. Appendix~\ref{sec:append-baseline} describes more details of baseline systems.

We describe our own model variants below,

\textbf{1.} \textsc{Vanilla-Gen} uses the parameters of a pretrained language model (LM), specifically BART-base \citep{lewis-etal-2020-bart}, to initialize both the storyline and story models. Its workflow is illustrated in the upper block of Figure~\ref{fig:model}. Since no information other than the prefix (first sentence, prompt, etc.) is used to generate the story, we denote this model as vanilla LM generation or \textsc{Vanilla-Gen}.

\textbf{2.} \textsc{Structured-Prompt} enhances \textsc{Vanilla-Gen} by using a structured storyline of events to encode \textbf{temporal prompts}, which is associated with the workflow of the bottom block of Figure~\ref{fig:model}. 

\textbf{3.} \textsc{Pretrained.} For ROCStories data only, we initialize the storyline model of \textsc{Structured-Prompt} with the pretrained parameters.

\textbf{4.} \textsc{RL} uses the same inputs as \textsc{Structured-Prompt}. The difference is that reinforcement learning is used to train storyline and story models jointly. As Algorithm~\ref{algo:1} shows, \textsc{RL}-based model is trained following the same forward workflow as \textsc{Structured-Prompt}, but during backpropagation, the storyline models' parameters are updated.

\section{Results and Analysis}
\label{sec:results}

\begin{table*}[t!]
\centering
\small
\scalebox{0.78}{
\setlength{\tabcolsep}{5pt}
\begin{tabular}{l|cccccc|ccc}
\toprule
 & \textbf{Ref.} & \textbf{Gen.} & \textbf{Distinct} & \textbf{BLEU} & \textbf{ROUGE}$_L$ & \textbf{Tok.} & \textbf{Pearson} & \textbf{Temporal} & \textbf{Interest}\\
\textbf{Models} & \textbf{PPL} ($\downarrow$) & \textbf{PPL} ($\downarrow$) & \textbf{Ratio} ($\uparrow$) & ($\uparrow$) & ($\uparrow$) & \textbf{Len.} ($\uparrow$) & \textbf{Corr.} ($\uparrow$) & \textbf{Coherence} ($\uparrow$) & \textbf{Level} ($\uparrow$)\\
\midrule
\textsc{ContentPlanning} & - & 25.52 & 1.80 & \textbf{3.46} & \textbf{14.40} & \textbf{252.3} & 0.04 & \textbf{0.57} & 2.20\\
\midrule
\textsc{Vanilla-Gen} & 31.04 & 11.17 & \textbf{3.50} & 0.67 & 9.43 & 160.2 & 0.09 & 0.52 & 2.49\\
\textsc{+ Structured-Prompt}& \textbf{30.77} & \textbf{9.30} & 2.86 & 1.44 & 10.95 & 208.6 & 0.56 & 0.49 & 2.49\\
\textsc{+ RL} & 30.98 & 9.50 & 2.83 & 1.39 & 10.78 & 203.8 & \textbf{0.57} & 0.55 & \textbf{2.62}\\
\bottomrule
\end{tabular}
}
\vspace{-0.3cm}
\caption{Evaluation results for WritingPrompts. Pearson correlation approximates the effectiveness of prompts.}
\vspace{-0.6cm}
\label{tab:res-wp}
\end{table*}

The main results for ROCStories and WritingPrompts are shown in Table~\ref{tab:res-roc} and Table~\ref{tab:res-wp} respectively. Examples of generated stories can be found in Table~\ref{tab:roc-examples} and Table~\ref{tab:roc-examples-more} for ROCStories and Table~\ref{tab:wp-examples-more} in the appendix for WritingPrompts. 
We organize our discussions and analysis in the following sections by answering the four research questions. \textbf{Q1)} Can our proposed models (with \textbf{temporal prompts}) produce stories with good \textbf{textual quality}? \textbf{Q2)} Are our proposed models effective at generating \textit{flashbacks}? \textbf{Q3)} Can our proposed models maintain event \textbf{temporal coherence} in stories? \textbf{Q4)} How do our proposed models contribute to stories' \textbf{interest levels}?

\subsection{Textual Quality}
We measure the textual quality of stories using a wide range of automatic evaluation metrics.

\paragraph{Perplexity.} For ROCStories, all three model variants can improve \textbf{Ref. PPL} against \textsc{Vanilla-Gen} and \textsc{Temporal-BART} while maintaining good \textbf{Gen. PPL}. The weak \textbf{Gen. PPL} of \textsc{MEGATRON} may be attributed to its sentence-by-sentence generation pipeline, whereas our models generate an entire story in an integrated step. For WritingPrompts, both model variants improve \textbf{Gen. PPL} over \textsc{Vanilla-Gen} and \textsc{ContentPlanning} while maintaining good \textbf{Ref. PPL}. 

\paragraph{Token Diversity.} For ROCStories, \textsc{RL}-based model improves the \textsc{Vanilla-Gen} by 0.18 per \textbf{Distinct Ratio}. \textsc{MEGATRON} achieves the highest token diversity as it incorporates external knowledge-bases that make the generated stories contain novel tokens. For WritingPrompts, we observe longer stories are associated with poorer scores. However, the large increases in \textbf{Distinct Ratio} suggest that the token usages in our proposed models are diverse. 

\paragraph{BLEU and ROUGE$_L$.} For ROCStories, the proposed models perform on-par with \textsc{Vanilla-Gen} and \textsc{Temporal-BART} while outperforming \textsc{MEGATRON}, which generates the shortest stories among all compared models. For WritingPrompts, \textsc{ContentPlanning} performs the best partially due to its usage of BART-large models.

The overall performances across these three types of automatic metrics suggest that using \textbf{temporal prompts} in the Plan-and-Write framework can produce stories with high textual quality.

\subsection{Effectiveness on Flashback Generation}
The second research question probes the effectiveness of using \textbf{temporal prompts} on generating \textit{flashbacks}. For ROCStories, all models can generate stories with the same number of events/sentences as the gold stories. This allows annotators to judge pairwise event relations in the generated stories and help us check whether the generated events have relations truthfully reflecting the \textbf{temporal prompts} used. \textbf{Accuracy} is the perfect metric for this. As Table~\ref{tab:res-roc} shows, the final \textsc{RL}-based model achieves the highest score, which indicates the strongest effectiveness of generating \textit{flashbacks}.

\begin{figure}[t!]
    \centering
\includegraphics[trim=0cm 0.0cm 0cm 0.0cm, width=0.8\columnwidth]{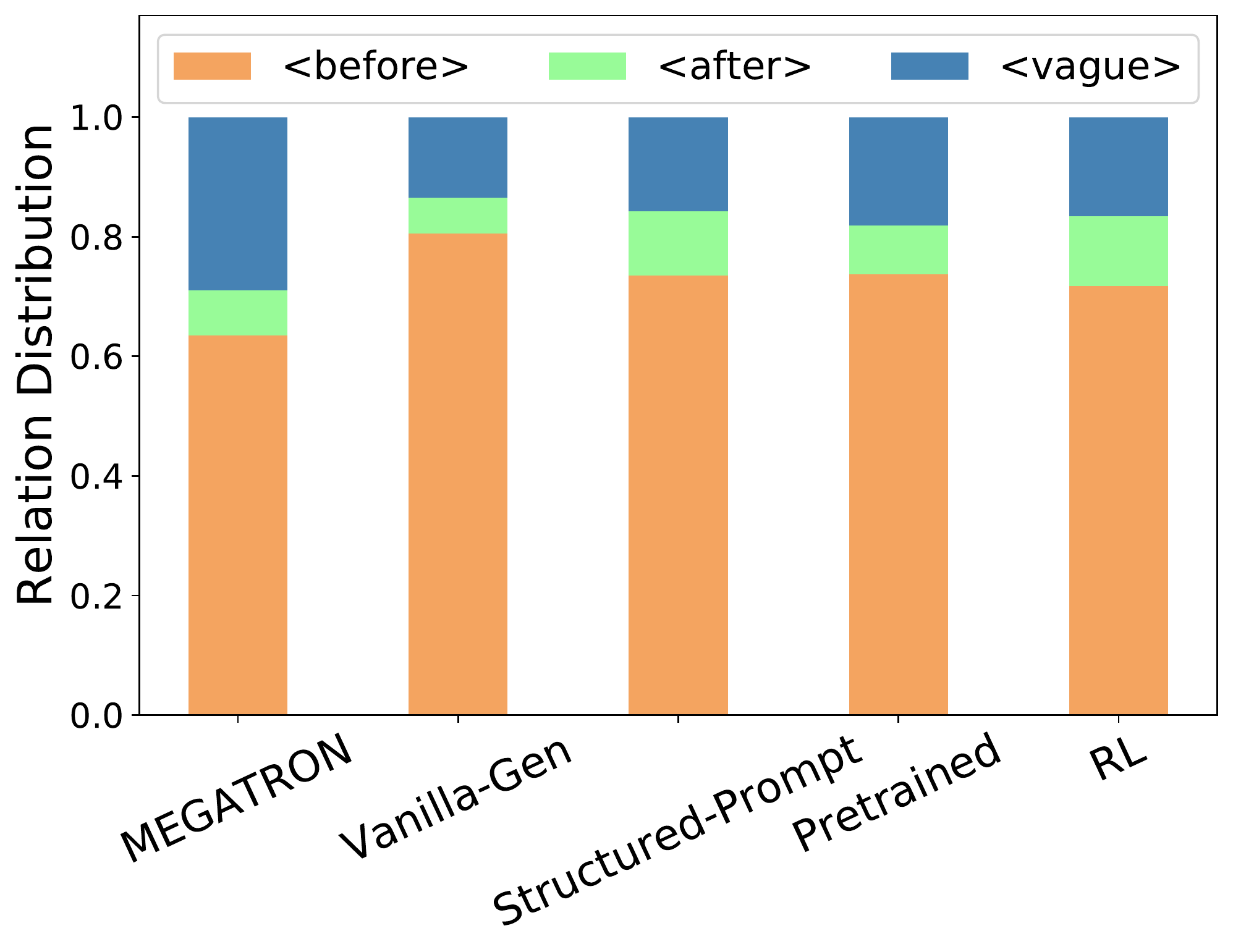}
\vspace{-0.5cm}
\caption{Relation annotation distribution by MTurkers for compared models trained on ROCStories.}
\label{fig:distribution}
\vspace{-0.5cm}
\end{figure}

However, \textbf{temporal prompts} are not used in the baselines and \textsc{Vanilla-Gen}. So we compute an approximate measures of effectiveness, \textbf{temporal diversity}, which indicates how many non-\temprel{before} relations \temprel{after} prompt can induce. Table~\ref{tab:res-roc} shows that \textsc{Structured-Prompt}, \textsc{Pretrained} and \textsc{RL}-based models can help improve \textsc{Vanilla-Gen} with more than 80\% generated \temprel{before} relations. \textsc{MEGATRON} achieves the highest score due to the largest amount (29\%) of \temprel{vague} relations (complex or undetermined) annotated by MTurkers shown in Figure~\ref{fig:distribution}, which is associated with its lowest \textbf{temporal coherence} score.

For WritingPrompts, stories are long and can contain dialogues or short phrases without events at all. These make the sentence or event alignments between the gold and generated stories worse than ROCStories, i.e. $e_{i,k}, r_{i,k}$ may not correspond to the k-th sentence in $\mathcal{Y}_i$. Therefore, \textbf{accuracy} cannot be computed. To obtain an approximate metric, we use the tool described in Sec.~\ref{sec:temp-prompt} to annotate neighboring event temporal relations in the \textit{generated} test stories for all the compared models. Slightly different from \textbf{temporal diversity}, we calculate the total number of machine annotated \temprel{after} relations, denoted as $\hat{N}_{i, A}$ in each $\hat{\mathcal{S}}_i'$. Let $\{N_{i, A}\}$ denote the number of \temprel{after} \textbf{temporal prompts} extracted in gold stories. We compute the \textbf{Pearson correlation} \citep{benesty2009pearson} between the sets $\{\hat{N}_{i, A}\}$ and $\{\hat{N}_{i, A}\}$ as the measure.

As Table~\ref{tab:res-wp} shows, for \textsc{ContentPlanning} and \textsc{Vanilla-Gen} without \textbf{temporal prompts}, the correlations are weak; whereas when \textbf{temporal prompts} are used in both \textsc{Structured-Prompt} and \textsc{RL}-based models, the correlations are strong. Although using models' temporal annotations for the generated stories is not as precise as human annotations, the large differences in correlation provide another piece of evidence that our proposed methods are effective at generating \textit{flashbacks}.

\subsection{Temporal Coherence}
Generating \textit{flashbacks} requires a system to disrupt the monotonic \temprel{before} sequence, which is the dominant temporal pattern generated by \textsc{Vanilla-Gen} (see Figure~\ref{fig:distribution}). In other words, \textit{flashbacks} with at least one \temprel{after} are minor patterns that can be hard to learn in our data, which may result in event sequences violating our temporal commonsense. Thus, we need to check that stories generated with \textit{flashbacks} maintain good \textbf{temporal coherence}. As shown in Table~\ref{tab:res-roc} and ~\ref{tab:res-wp}, our proposed models with \textbf{temporal prompts} can achieve on-par or slightly lower scores, suggesting little trade-off of \textbf{temporal coherence} in generating \textit{flashbacks}. We will discuss this more in the error analysis (Sec.~\ref{sec:error}).

\subsection{Contributions to the Interest Level}
As we can observe in Table~\ref{tab:res-roc}, the impact of \textbf{temporal diversity} and \textbf{coherence} on the \textbf{interest level} appears to be complex. To better understand the dynamics among these metrics, we run ordinary least square regressions (OLS) \cite{OLS} by setting \textbf{interest level} as the target variable and \textbf{temporal diversity}, \textbf{coherence} and the number of \temprel{after}s as predictors. Since all of these metrics apply to each of the compared stories, the total instances are 530 and 308 for ROCStories and WritingPrompts, respectively.

As Table~\ref{tab:statistical-test} shows, for ROCStories, holding other metrics constant, adding 1 unit to \textbf{temporal coherence} and \textbf{diversity} leads to a 0.609 increase and 0.532 decrease of the \textbf{interest level}. The former result implies that a story lacking event temporal coherence tends to be less interesting. The latter result suggests that increasing \textbf{temporal diversity} may lead to less interesting stories, which we hypothesize could be attributed to two factors: 1) \temprel{before} is dominant in ROCStories, and by using \temprel{after} as prompt, we force models to generate relations less seen in data. 2) Figure~\ref{fig:distribution} shows \textbf{temporal diversity} can increase with more \temprel{vague} relations. Since \temprel{vague} is an undetermined temporal relation even for our annotators, it could make the storyline confusing and thus lead to less interesting stories. The coefficient for the number of \temprel{after} indicators is positive with strong statistical significance. It suggests that holding the other two metrics constant, adding the number of \temprel{after} indicators by 1 contributes to 0.387 increases of the \textbf{interest level}.

For WritingPrompts, although we are not able to conclude that the estimates are statistically significant, the coefficients have the same signs as ROCStories. Also, we observe that the p-value of the number of \temprel{after} indicators is much lower than the other two variables, which implies a relatively stronger (positive) impact.

Since \textbf{temporal prompts} in human evaluations all contain at least one \temprel{after}, these results show that when \temprel{after} prompt successfully produces event pairs with \temprel{after} relation in the final stories, it makes stories more interesting. Now, we can answer the final research question: improving \textbf{temporal diversity} can help \textbf{interest level} when \temprel{after} prompts are effective at generating \temprel{after} relations in stories; that is, when \textit{flashbacks} truly work, stories become more interesting. 
 
 \begin{table}[t]
\centering
\small
\scalebox{0.8}{
\setlength{\tabcolsep}{6pt}
\begin{tabular}{l|cc|cc}
\toprule
 & \multicolumn{2}{c|}{\textbf{ROCStories}} & \multicolumn{2}{c}{\textbf{WritingPrompts}} \\
 \midrule
 & \textbf{Coef.} & \textbf{p-value} & \textbf{Coef.} & \textbf{p-value}\\
\midrule
\textbf{Temporal Coherence} & 0.609 & 0.000$^{*}$ & 0.006 & 0.963 \\
\textbf{Temporal Diversity} & $-$0.532 & 0.004$^*$ & $-$0.279 & 0.410 \\
\textbf{\# \temprel{after} prompt} & 0.387 & 0.000$^{*}$ & 0.034 & 0.238\\
\bottomrule
\end{tabular}
}
\vspace{-0.3cm}
\caption{OLS regression results on temporal coherence, diversity and number of \temprel{after} indicators. The coefficients for the intercept are omitted. $^{*}$ means statistically significant with 99\% confidence.}
\label{tab:statistical-test}
\vspace{-0.6cm}
\end{table}

\section{Error Analysis}
\label{sec:error}
In Table~\ref{tab:res-roc}, we observe that our final models cannot outperform \textsc{VANILLA-GEN} for \textbf{Temporal Coherence} and fall behind \textsc{MEGATRON} for \textbf{Temporal Diversity}. We show examples below to provide feasible explanations.
\vspace{-0.2cm}
\paragraph{Temporal Coherence.} Table~\ref{tab:error-coherence} shows three comparisons between our final \textsc{RL}-based model and the \textsc{VANILLA-GEN} baseline. In all three pairs, human evaluators mark \textsc{VANILLA-GEN}'s generated stories as temporally coherent but not for \textsc{RL}'s stories. As we can see in these examples, the red highlighted events temporally contradict the preceding context. In Input 1, the context shows ``I have listened to the new album'' so ``can't wait to listen to the new album'' shouldn't follow the former event. For Input 2, if ``Anna had gotten her hair permed,'' it does make sense that the event ``went to apply it (the perm)'' follows. In Input 3, if the officer ``gave [a] ticket'' already, it does quite make sense to give ``a warning'' afterward as it is a weaker penalty than a ``ticket.'' However, we observe that in all three cases, \temprel{after} relations are successfully reflected (\textit{flashback}). Since stories with \temprel{after} relations are minority cases in the data, we hypothesize that they have not been perfectly learned by our proposed models. We leave more rigorous investigation for future research efforts.

 \begin{table}[t]
    \small
    \setlength{\tabcolsep}{1pt}
    \scalebox{0.8}{
     	\centering
 	\setlength\extrarowheight{1pt}
 	\begin{tabular}{ll} \hline
 	\toprule
 	\textbf{Input 1} & [s1] the new album from my favorite pop star came out today. \\
 	\textbf{prompt} & \textbf{\temprel{after}} \temprel{before} \temprel{before} \temprel{before}  \\
 	\midrule
 	\rowcolor{light-gray}  & [s2] i was so excited to listen to it. \\
 	\rowcolor{light-gray} & [s3]  i listened to it all day long.   \\ 
 	\rowcolor{light-gray} \multirow{-2}{*}{\textsc{VANILLA}} & [s4] i couldn't believe how good it was.  \\
 	\rowcolor{light-gray} & [s5] red i can't wait to listen again. \\ 
 	  & [s2] \hl{i had never heard the album before.} \\
 	 & [s3] i went to the store and bought the album.  \\ 
 	 \multirow{-2}{*}{\textsc{RL}} & [s4] when i got home, i listened to the whole thing.  \\
 	 & [s5] \color{red}i can't wait to listen to the new album.  \\ 
 	
    \toprule
 	\textbf{Input 2} & [s1] anna had just gotten her hair permed. \\
 	\textbf{prompt} &  \temprel{before}  \temprel{before} \textbf{\temprel{after}} \temprel{before}  \\
 	\midrule
 	\rowcolor{light-gray}  & [s2] she was excited to see how it looked. \\
 	\rowcolor{light-gray} & [s3] but then she saw that her hair had grown out.  \\ 
 	\rowcolor{light-gray} \multirow{-2}{*}{\textsc{VANILLA}} & [s4] anna was horrified!  \\
 	\rowcolor{light-gray} & [s5] she had to go to the salon to get it fixed. \\ 
 	  & [s2] she was excited to see how it looked. \\
 	 & [s3] \color{red} but when she went to apply it, it was terrible.   \\ 
 	 \multirow{-2}{*}{\textsc{RL}} & [s4] \hl{her hair had grown way too long!}  \\
 	 & [s5] anna had to go to the salon to get her hair fixed. \\ 
 	 
    \toprule
 	\textbf{Input 3} & [s1] last week we saw a police car in front of a parked car. \\
 	\textbf{prompt} & \textbf{\temprel{after}} \temprel{before}  \temprel{before} \temprel{before}  \\
 	\midrule
 	\rowcolor{light-gray}  & [s2] the police officer pulled over the car. \\
 	\rowcolor{light-gray} & [s3] he asked the driver if he had insurance.   \\ 
 	\rowcolor{light-gray} \multirow{-2}{*}{\textsc{VANILLA}} & [s4] the driver said he did not.  \\
 	\rowcolor{light-gray} & [s5] the officer gave him a ticket. \\ 
 	  & [s2] \hl{the car was parked in the parking lot.} \\
 	 & [s3] the police officer pulled over the car.   \\ 
 	 \multirow{-2}{*}{\textsc{RL}} & [s4] he gave the driver a ticket.  \\
 	 & [s5] \color{red} the officer gave him a warning. \\ 
 	\bottomrule
 	\end{tabular}
 	}
 	\vspace{-0.2cm}
   	\caption{Examples from our \textsc{RL}-based model that are not as temporally coherent as examples generated by the \textsc{Vanilla-Gen} model. Highlighted sentences contain an event that started earlier than the preceding event. Red text indicates temporal incoherence.}
   	\label{tab:error-coherence}
   \vspace{-0.7cm}
 \end{table}
 \vspace{-0.1cm}
 \paragraph{Temporal Diversity.} 
Table~\ref{tab:error-diversity} in the appendix shows three comparisons between our final \textsc{RL}-based model and the \textsc{MEGATRON} baseline. In all three pairs, \textsc{MEGATRON}'s generated stories are more temporally diverse based on the \textbf{predicted relations} provided by human evaluators. However, \textsc{MEGATRON}'s stories are either contradictory (Input 1), incoherent (Input 2) or repetitive (Input 3), resulting in higher ambiguous event relations, i.e. \temprel{vague} annotations (consistent with Figure~\ref{fig:distribution}). Therefore, despite lower temporal diversity, our proposed models can still produce stories with higher quality, which is demonstrated via other metrics such as \textbf{Temporal Coherence} and \textbf{Interest Level}.

\section{Related Work}
\label{sec:related}
\paragraph{Generating flashbacks} has been studied in a few prior works. \citet{10.1007/978-3-540-89454-4_22} is one of the early efforts proposing a planning-based approach to generate flashbacks to evoke surprise in the readers. Follow-up works proposed a cognitive-based model that finds the best location in the original stories to insert a past event \citep{wu:hal-01413401}. Our work differs from this line of research by using temporal prompts with pretrained language models to generate integrated \textit{flashback} in stories. \citet{Hoek_Theune_Linssen_2014} studies \textit{flashback} in game narrative generation, which is remotely related to our work.
\vspace{-0.2cm}
\paragraph{Plan-and-Write framework} has been shown to be an effective method to enhance the explainability and controllability of story generation. \citet{yao2019plan} enables machines to produce a sequence of keywords prior to generating stories. Follow-up works leverage commonsense or external knowledge to enhance the quality of stories \citep{guan-etal-2020-knowledge, xu-etal-2020-megatron, tanprogressive}.
\citet{goldfarb-tarrant-etal-2020-content} is one of our compared works that incorporates SRL extracted event representations in storylines and train models with several event-related decoding objectives. Our work differs from it by explicitly encoding \textbf{temporal prompts} in event plots that facilitates \textit{flashback}.
\vspace{-0.2cm}
\paragraph{Structured representation} such as discourse structure \citep{guan-etal-2021-long}, story keywords~\citep{peng2018towards,goldfarb-tarrant-etal-2019-plan} and event/plot graph \citep{ammanabroluaaai2020,ammanabrolu2021automated} have been widely used in story generation to enable models to output diverse stories, but they are remotely related to our \textit{flashback} generation task.
\vspace{-0.2cm}
\paragraph{Reinforcement learning} has also been explored in two-stage story generation such as \citet{xu-etal-2018-skeleton} and \citet{tambwekarijcai2019}. Our motivation of using RL-based generation is to enhance the effectiveness of \textbf{temporal prompts}.
\vspace{-0.2cm}
\paragraph{Event temporal reasoning} helps the construction of the \textbf{temporal prompts}. It has been studied in story understanding~\cite{han2019contextualized}, information extraction \cite{ning-etal-2017-structured, han-etal-2019-joint, han-etal-2019-deep, han-etal-2020-domain, ma2021eventplus}, QA \citep{ning-etal-2020-torque, zhou-etal-2019-going}, and event generation \cite{lin-etal-2021-conditional, li2021future}. 
Our proposed method is the pioneer work to introduce event \textbf{temporal prompts} in \textit{flashback} generation.
\section{Conclusions}
We propose to generate \textit{flashbacks} in stories by encoding temporal prompts in structured storylines. Experimental and evaluation results show our proposed systems can produce fluent, temporally coherent, and more interesting stories. Future work can focus on improving temporal prompts so that they handle minority cases (i.e. prompts with one or more \temprel{after} indicators) more effectively. Designing more powerful rewards in reinforcement learning is another promising research direction.
\section*{Acknowledgments}
We thank the PlusLab members and the anonymous reviewers for their constructive feedback. This work is supported in part by IARPA, via Contract No. 2019-19051600007, the DARPA Machine Common Sense (MCS) program under Cooperative Agreement N66001-19-2-4032, and a CISCO research award. 

\bibliographystyle{acl_natbib}
\bibliography{anthology,acl2022}

\clearpage
\appendix

\section{Additional Generated Stories}
\label{sec:roc-more-exp}
Please see Table~\ref{tab:roc-examples-more} and Table~\ref{tab:wp-examples-more} for more examples on ROCStories and WritingPrompt respectively.

\begin{table}[t]
    \small
    \setlength{\tabcolsep}{1pt}
    \scalebox{0.8}{
 	\centering
 	\setlength\extrarowheight{1pt}
 	\begin{tabular}{ll} \hline
 	\toprule
 	\textbf{Input} & [s1] nina needed blood tests done. \\
 	\textbf{prompt} & \temprel{before} \textbf{\temprel{after}} \temprel{before} \temprel{before}  \\
 	\midrule
 	\rowcolor{light-gray}  & [s2] she was very nervous. \\
 	\rowcolor{light-gray} & [s3]  she went to the local clinic.   \\ 
 	\rowcolor{light-gray} \multirow{-2}{*}{\textbf{baseline}} & [s4] they told her about her bad breath.  \\
 	\rowcolor{light-gray} & [s5] she was relieved her nerves were n't too much to handle. \\ 
      & [s2] she was terrified of needles. \\
 	\multirow{2}{*}{\textsc{Vanilla}} & [s3] she went to the doctor. \\
      & [s4] the doctor gave her a needle. \\
     & [s5] nina was relieved.\\
 	\rowcolor{light-gray}  & [s2] she was afraid to go to the hospital. \\
 	\rowcolor{light-gray} & [s3] \hl{she had never been to a hospital before.}   \\ 
 	\rowcolor{light-gray} \multirow{-2}{*}{\textsc{RL}} & [s4] but when she got there, she was relieved.  \\
 	\rowcolor{light-gray} & [s5] she felt so relieved!   \\ 
 	\bottomrule \\
 	
    \toprule
 	\textbf{Input} & [s1] mickey was always a fan of webcomics. \\
 	\textbf{prompt} &  \textbf{\temprel{after}} \temprel{before}  \temprel{before} \temprel{before}  \\
 	\midrule
 	\rowcolor{light-gray}  & [s2] he loved writing his own comics one night. \\
 	\rowcolor{light-gray} & [s3] then he had a great idea.   \\ 
 	\rowcolor{light-gray} \multirow{-2}{*}{\textbf{baseline}} & [s4] he created a story called the superman story.   \\
 	\rowcolor{light-gray} & [s5] the story was real as a kids cartoon. \\ 
      & [s2] one day, he decided to make a webcomic himself. \\
 	\multirow{2}{*}{\textsc{Vanilla}} & [s3] he created a website and sent it to a publisher. \\
      & [s4] the publisher loved it and contacted mickey. \\
     & [s5] mickey was happy to receive a publishing deal.\\
 	\rowcolor{light-gray}  & [s2] \hl{he had been writing for years.} \\
 	\rowcolor{light-gray} & [s3] one day, mickey saw an ad for a webcomic.   \\ 
 	\rowcolor{light-gray} \multirow{-2}{*}{\textsc{RL}} & [s4] he signed up for it.  \\
 	\rowcolor{light-gray} & [s5] mickey was able to make a lot of money from his work. \\ 
 	\bottomrule \\
 	
    \toprule
 	\textbf{Input} & [s1] anna had just gotten her hair permed. \\
 	\textbf{prompt} & \temprel{before} \temprel{before}   \textbf{\temprel{after}} \temprel{before}  \\
 	\midrule
 	\rowcolor{light-gray}  & [s2] she was nervous about it. \\
 	\rowcolor{light-gray} & [s3] she was too nervous to do it.   \\ 
 	\rowcolor{light-gray} \multirow{-2}{*}{\textbf{baseline}} & [s4] she tried to cut hair to look better.   \\
 	\rowcolor{light-gray} & [s5] when she was done it looked great !  \\ 
      & [s2] she was excited to see how it looked. \\
 	\multirow{2}{*}{\textsc{Vanilla}} & [s3] but then she saw that her hair had grown out. \\
      & [s4] anna was horrified!  \\
     & [s5] she had to go to the salon to get it fixed.\\
 	\rowcolor{light-gray}  & [s2] she was excited to see how it looked. \\
 	\rowcolor{light-gray} & [s3] but when she went to apply it, it was terrible.   \\ 
 	\rowcolor{light-gray} \multirow{-2}{*}{\textsc{RL}} & [s4] \hl{her hair had grown way too long!} \\
 	\rowcolor{light-gray} & [s5] anna had to go to the salon to get her hair fixed. \\ 
 	\bottomrule
 	\end{tabular}
 	}
   	\caption{Additional generated stories for ROCStories.}
   	\label{tab:roc-examples-more}
 \end{table}

\section{Data}
\label{sec:append-data}
\paragraph{ROCStories}\citep{mostafazadeh-etal-2016-corpus}  contains 5-sentence stories. Following \citep{xu-etal-2020-megatron}, we split data into 88,344/4,908/4,909 for train/validation/test sets. 

\paragraph{WritingPrompt}\cite{fan-etal-2018-hierarchical} contains 30,335 pairs of prompts and stories. With an average of more than 700 words per story, Writing Prompts are much longer than ROCStories. These stories are also much less structured as some dialogues and short phrases may be included. To speed up our experiments, we select stories with a maximum of 500 words, resulting in a total number of 96,488 training and 5,784 validation prompt-story pairs, respectively. For the test set, we use the 1,000 prompt-story pairs provided by the baseline paper \cite{goldfarb-tarrant-etal-2020-content} for reporting automatic evaluation results.

\paragraph{Pretraining  Data.} As we mention in Section~\ref{sec:pretrain}, we pretrain storyline models for ROCStories. To be consistent with ROCStories inputs, we divide BookCorpus data \citep{Zhu_2015_ICCV} into 5 consecutive sentences and filter out those with noisy tokens. We randomly select 1 million such 5-sentence text spans and extract their storylines following Section~\ref{sec:struct-input}.

\section{More Details for Evaluation Metrics}

\paragraph{Automatic evaluation metrics} are used to measure textual quality of stories. We report 1) \textbf{Ref. PPL}: reference stories' perplexity in a model; 2) \textbf{Gen. PPL}: generated stories' perplexity scored by GPT-2 \citep{radford2019language}, i.e. we feed the generated stories into GPT-2 to compute perplexity scores. For diversity scores, we found our models implemented by Huggingface \citep{wolf-etal-2020-transformers} can achieve nearly 0 \textbf{Repeat-3} and 100\% \textbf{Distinct-3} scores, so we follow \citet{goldfarb-tarrant-etal-2020-content} to compute the overall vocabulary:token number ratio, which we denote as 3) \textbf{Distinct Ratio} (\%). We also report standard 4) \textbf{BLEU-3} and 5) \textbf{ROUGE}$_L$ scores.

\section{More Details for Baseline Models}
\label{sec:append-baseline}
\paragraph{MEGATRON-CNTRL} \citet{xu-etal-2020-megatron}, denoted as \textsc{MEGATRON} for brevity, is chosen as the baseline as it outperforms previous systems such as \citet{guan-etal-2020-knowledge} on ROCStories. We do not perform delexicalization that replaces names and entities with \textit{[MALE]}, \textit{[FEMALE]} and \textit{[NEUTRAL]} tokens, as we found our models work well by recognizing names and entities. When conducting evaluations, we try our best to \textbf{map these special tokens back to their original texts} by using the given first sentence. For rare undetermined cases, we manually examine the generated stories and swap in names or entities that make the most sense in the context. To be fair, we compare with the 124M-parameter version.

\paragraph{ContentPlanning} \citep{goldfarb-tarrant-etal-2020-content} is chosen as the baseline for WritingPrompt, as it also adopts the Plan-and-Write workflow as well as structured event representations. However, their models are based on BART-large and do not train with an end-to-end framework. They use 65\% of the original training data and also filter out samples with non-[WP] prompts. Our final training data is about 2/3 of theirs.

\paragraph{TemporalBART} \citep{lin-etal-2021-conditional} is designed for two event temporal relation related tasks: temporal ordering and event infilling. Although TemporalBART does not tackle story generation directly, it encodes event temporal information via pretraining tasks. So we consider TemporalBART as another baseline model by initialing the storyline model with their parameters and training the \textsc{Structured-Prompt} workflow on ROCStories.\footnote{More implementation details such as hyper-parameters and software can be found in the appendix.}

\section{Reproduction Check List}
We finetune BART-base. For ROCStories, hyper-parameters are learning rate: $ 5e^{-5}$; batch size: $10$. We use 3 random seeds: $(5, 9998, 20016)$ and report the average performances for all end-to-end models. For Writing Prompts, hyper-parameters are learning rate: $ 1e^{-4}$; batch size: $64$; gradient accumulation: $8$.

For ROCStories, we were able to finetune on a single Nvidia GTX2020 GPU with 11G memory, and training time is 3-4 hours per epoch. For WritingPrompt, we have to use a much larger Nvidia A100 GPU with 40G memory, and the training time is 20 hours per epoch. We train all models for 10 epochs and save the model with the best evaluation perplexity. All reproduction details can be found in the separately submitted code.
 
\section{Perplexity and Event Coverage Trade-off}
\label{sec:tradeoff}
One caveat of using end-to-end training is that there is no guarantee that the generated events will appear in the final stories;
whereas in two-stage models, the story model learns a mapping from reference storylines to stories, which leads to a higher coverage rate of the generated events. To provide a potential solution, we experiment with the mixture-training method proposed by \citet{zhang-etal-2019-bridging}, 
\[
p = \frac{\mu}{\mu + \exp(e/\mu)}
\]
where p controls the ratio of reference storylines used in training and $e$ is the training step. Here the larger the hyper-parameter $\mu$, the slower p decays to 0 as training proceeds.

\begin{figure}[t!]
    \centering
\includegraphics[width=\columnwidth]{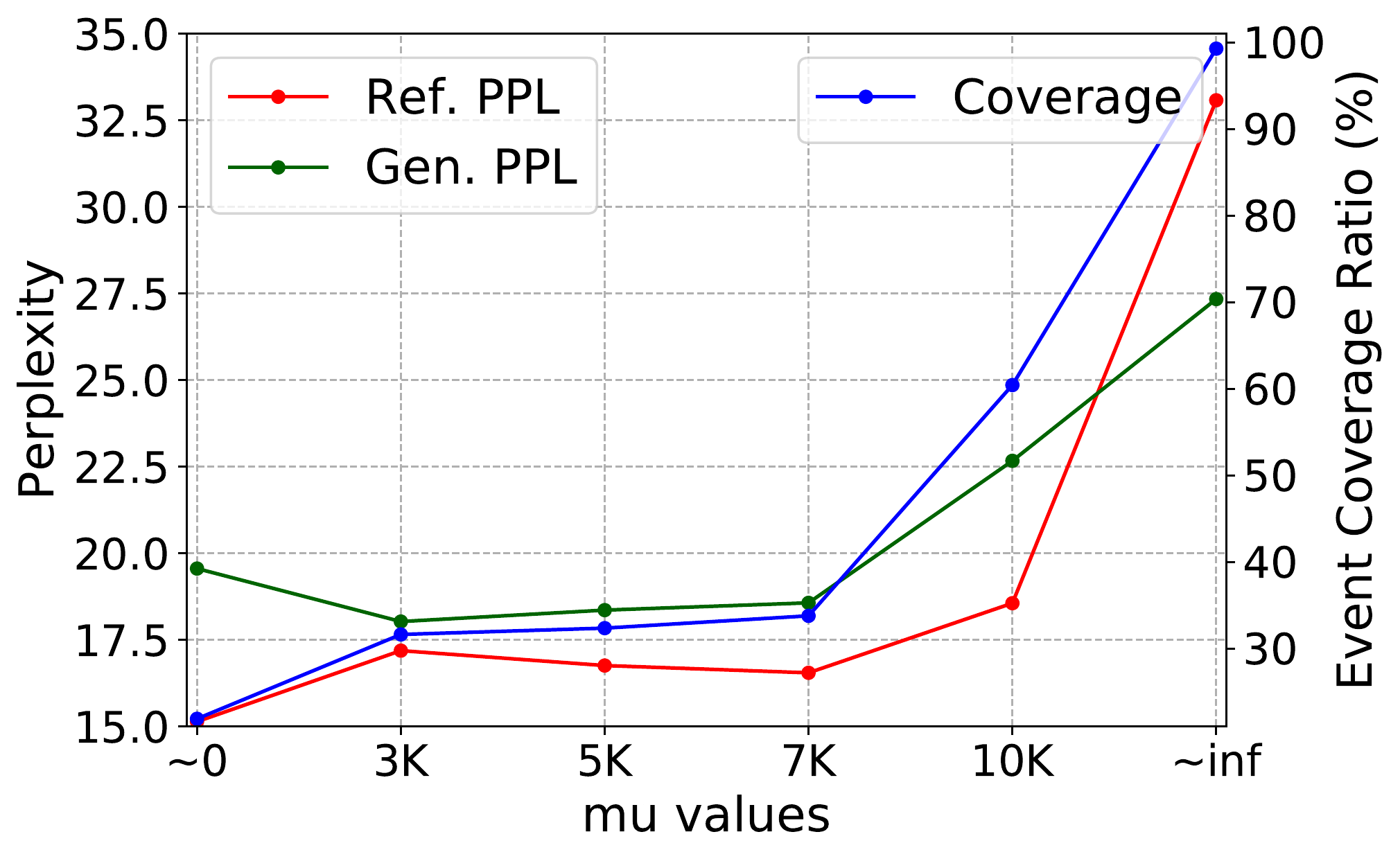}
\caption{Trade-off between perplexity scores and generated event (trigger) coverage.}
\label{fig:ppl-coverage}
\end{figure}

In Figure~\ref{fig:ppl-coverage}, we show the trade-off between perplexity and $\mu$ values by training our final RL-based models. When $\mu$ is nearly zero, it corresponds to always using predicted storylines, and hence a smaller predicted event coverage rate in the final stories. When $\mu$ gets larger, eventually making p close to 1, the training corresponds to always using gold storylines, which leads to a very high event coverage rate, but relatively poor perplexity (still much stronger than two-stage results). We leave the search for optimal $\mu$ for future research.

\begin{table}[h!]
\centering
\scalebox{0.8}{
\setlength{\tabcolsep}{5pt}
\begin{tabular}{l|l}
\toprule
CaTeRS & MATRES \\
\midrule
A \temprel{Before} B & A \temprel{before} B\\
A \temprel{Identity} B & A \temprel{vague} B\\
A \temprel{Contains} B & A \temprel{before} B\\
A \temprel{Overlaps} B & A \temprel{before}, \temprel{after}, \temprel{vague} B\\
\bottomrule
\end{tabular}
}
\caption{Label mapping from CaTeRS to MATRES. \temprel{after} is ignored in CaTeRS by flipping event physical order.}
\label{tab:caters-matres}
\end{table}

\section{Benchmark Event Temporal Relation Annotations}
\label{sec:benchmark}
The experimental results in the main text demonstrate the effectiveness of using temporal prompts. Here, we further show that the tool to produce temporal prompts, i.e. ECONET, provides reliable event temporal relation annotations. We benchmark ECONET's performances using CaTeRS \cite{mostafazadeh-etal-2016-caters}, which annotates 4 types of temporal relations for event pairs in a small amount of ROCStories. However, CaTeRS's annotations are based on event time interval rather than event start time as used in MATRES, which ECONET is finetuned on.

In Tabel~\ref{tab:caters-matres}, we provide a mapping from CaTeRS's temporal relations to MATRES labels.  The only non-unique mapping is \temprel{Overlap}. In other words, when ECONET predicts \temprel{before} for a CaTeRS sample \temprel{Overlaps}, we have to manually examine whether it is correct or not. We found that when ECONET predicts \temprel{before} for CeTeRS data, the precision rate is 65.53\% due to a large amount of \temprel{Overlaps} event pairs being predicted as \temprel{before}. But we emphasize here that this low number \textbf{is caused by label mismatch} as shown in Table~\ref{tab:caters-matres}, which does not truthfully reflect the ECONET's accuracy.

To have a better understanding, we randomly selected 20 such pairs and manually examine their temporal relations in the context and found that 90\% of such pairs are indeed correctly predicted by ECONET. Adjusting for this factor, the precision rate for the annotated \temprel{before} relation would be 92.07\%, indicating highly accurate predictions. We do not claim the final accuracy is 92.07\%, but simply argue that the annotations provided by ECONET are helpful as our main experimental results demonstrate.

\section{Two-stage Model Results}
\paragraph{Two-stage Model.} As we mentioned in Section~\ref{sec:method}, another way to implement Plan-and-Write framework is to train storyline and story models separately with gold input and outputs, and replace story models' inputs with the storyline model's predictions during inference. We found this variant's performances fall far behind other compared models. So we do not use them in human evaluations and simply show their automatic evaluation results in Table~\ref{tab:res-two-stage}.

\begin{table}[htbp!]
\centering
\small
\scalebox{0.8}{
\setlength{\tabcolsep}{4pt}
\begin{tabular}{l|ccccc}
\toprule
 & \textbf{Ref.} & \textbf{Gen.} & \textbf{Distinct} & \textbf{BLEU} & \textbf{ROUGE}$_L$ \\
\textbf{Models} & \textbf{PPL}($\downarrow$) & \textbf{PPL}($\downarrow$) & \textbf{Ratio}($\uparrow$) &  ($\uparrow$) &  ($\uparrow$) \\
\midrule
ROCStories & 197.97 & 27.15 & 4.26 & 2.98 & 16.39 \\
\midrule
WritingPrompt & 76.69 & 39.43 & 8.12 & 0.00 & 6.24 \\
\bottomrule
\end{tabular}
}
\caption{Automatic Evaluation Results for the Two-stage models.}
\label{tab:res-two-stage}
\end{table}

\section{Storyline Model Results}
Our primary goal is to improve temporal control in final stories. However, as aforementioned, the motivation for using reinforcement learning is to allow the storyline model to adapt together with the story model. In Table~\ref{tab:res-storyline}, we compare storyline prediction performances between the vanilla and the RL-based end-to-end models. These results suggest that our final framework combining structured storylines, temporal prompts, pretraining and RL also helps storyline models fit reference storylines better, resulting in lower perplexity, higher overlapped scores and better diversity.

\begin{table}[t!]\vspace{-0.3cm}
\centering
\small
\scalebox{0.8}{
\setlength{\tabcolsep}{5pt}
\begin{tabular}{l|ccc}
\toprule
 & \textbf{Ref.} & \textbf{BLEU} & \textbf{Distinct}\\
\textbf{Models} & PPL ($\downarrow$) &  ($\uparrow$) & \textbf{Ratio} ($\uparrow$)\\
\midrule
\textbf{Vanilla-Gen} & 13.97 & 3.21 & 3.68 \\
\textbf{+ RL} & \textbf{12.71} & \textbf{3.49} & \textbf{4.09} \\
\bottomrule
\end{tabular}
}
\vspace{-0.2cm}
\caption{Improvements of RL-based model against vanilla end-to-end model on ROCStories data.}
\label{tab:res-storyline}
\vspace{-0.5cm}
\end{table}

\section{Error Analysis}
Table~\ref{tab:error-diversity} shows examples for the temporal diversity in Section~\ref{sec:error}.
\begin{table}[h!]
    \small
    \setlength{\tabcolsep}{1pt}
    \scalebox{0.78}{
 	\centering
 	\setlength\extrarowheight{1pt}
 	\begin{tabular}{ll} \hline
 	\toprule
 	\textbf{Input 1} & [s1] justin is a very charismatic man who loves women. \\
 	\midrule
 	  & [s2] \color{red} he always said that he wanted to be a man. \\
 	 & [s3]  \color{red} he thought that men looked bad.   \\ 
 \multirow{-2}{*}{\textbf{baseline}}	 & [s4] justin decided to become a boy scout.  \\
 	 & [s5] he is now a successful realtor. \\ 
 & \textbf{predicted relations:} \temprel{vague} \temprel{vague} \temprel{vague}  \temprel{before} \\
 	 \midrule
 	  & [s2] one day he met a beautiful woman and they fell in love. \\
 	 & [s3] justin decided he wanted to marry her.   \\ 
 \multirow{-2}{*}{\textsc{RL}} 	  & [s4] he went to the mall and bought a ring.  \\
 	 & [s5] he proposed to her and she said yes.  \\ 
  & \textbf{predicted relations:} \temprel{before} \temprel{before}  \temprel{before}  \temprel{before} \\
    \toprule
 	\textbf{Input 2} & [s1] the smoke was beginning to get to me. \\
 	\midrule
 	  & [s2] mary friend said he had to come to my house. \\
 	 & [s3]  \color{red}i wanted to have a smoke fight.  \\ 
 \multirow{-2}{*}{\textbf{baseline}}	 & [s4] \color{red}he said i had to light a cigarette.  \\
 	 & [s5] i smoked for a few hours before i gave up. \\ 
 & \textbf{predicted relations:} \temprel{after} \temprel{vague} \temprel{before}  \temprel{before} \\
 	 \midrule
 	  & [s2] i had just gotten out of my car. \\
 	 & [s3] i ran to the front door.    \\ 
 \multirow{-2}{*}{\textsc{RL}} 	  & [s4] i opened the door. \\
 	 & [s5] the fire department came and put out the fire.  \\ 
  & \textbf{predicted relations:} \temprel{vague} \temprel{before}  \temprel{before}  \temprel{before} \\
 	 
    \toprule
 	\textbf{Input 3} & [s1] tom was in court for a parking ticket. \\
 	\midrule
 	  & [s2] he did not have to pay his court date. \\
 	 & [s3]  \color{red} the judge gave him a parking ticket.   \\ 
 \multirow{-2}{*}{\textbf{baseline}}	 & [s4] \color{red} tom paid his court date. \\
 	 & [s5] \color{red} the judge gave tom a parking ticket. \\ 
 & \textbf{predicted relations:} \temprel{before} \temprel{vague} \temprel{before}  \temprel{after} \\
 	 \midrule
 	  & [s2] he had been speeding for a while. \\
 	 & [s3] the judge gave him a warning.   \\ 
 \multirow{-2}{*}{\textsc{RL}} 	  & [s4] tom didn't listen to the warning.  \\
 	 & [s5] he got a ticket anyway.  \\ 
  & \textbf{predicted relations:} \temprel{after} \temprel{before}  \temprel{before}  \temprel{after} \\
 	\bottomrule
 	\end{tabular}
 	}
 	\vspace{-0.2cm}
   	\caption{Examples from our \textsc{RL}-based model that are not as temporally diverse as examples generated by the baseline \textsc{MEGATRON} model. \textbf{Predicted relations} are provided by human evaluators.}
   	\label{tab:error-diversity}
 \end{table}

\section{Potential Risks}
Since our models deal with open-domain generation, it is conceivable that the generated stories could contain biases, malicious languages, and hallucinations. We refer readers to the work in fairness and fact checking to address these issues.

 \begin{table*}[t]
    \small
    \setlength{\tabcolsep}{1pt}
    \scalebox{0.8}{
 	\centering
 	\setlength\extrarowheight{1pt}
 	\begin{tabular}{lp{18cm}} \hline
 	\toprule
 	\textbf{Input} &  artic outpost 9\\     
 	\textbf{prompt} &  \temprel{after} \temprel{vague} \temprel{before} \temprel{after} \temprel{before} \temprel{vague} \temprel{vague} \\
 	\midrule
 	\rowcolor{light-gray} \multirow{-1}{*}{\textbf{baseline}} &  she stood alone in the dimly lit corridor , staring at the light that flooded into the room . her eyes scanned the room , finally settling on the small , round table . <p> <p> `` hello , everyone . '' <p> the woman spoke in a voice that echoed the room in a series of hushed tones , as though it was a voice only audible to those who heard it . `` do you know why we are here ? do you know where we come from ? '' our leader , the man with the black suit and black hat , stood at the end of the hall . a man in a black coat and black trousers , a long coat , and a black hat stood behind him , a black cloak and a dark hat , a pair of gloves . the black suit had a black mask on , and she was wearing her best , black dress shirt and a pairof pants , and black shoes . on her right wrist , the small of one shoulder , was an amulet , and on the other , a small device hidden in the pocket of the man 's coat , a single small device . there was nothing else . just this man . . . and our leader\\

      \multirow{-1}{*}{\textsc{Vanilla}} & i've never seen anything like it before. it's a strange place. i don't know why i chose it, but i've always wanted to see it.  it was the first time i saw it. it was a small building with a single window. i couldn't believe it. i had never seen it before, but it was here.  i'm not sure what it is, but there's something about it that makes me want to run. i can't remember the last time i 'd seen it. but i do know that it was there.  the first thing i noticed was the door opening. i wasn't sure if it was my own, or my own. i didn't want to be here. i just wanted to get out. i wanted to go back to my home.  as i approached the door, i saw a man standing there. he was wearing a trench coat. he had a gun pointed at his head. he pointed it at my head.  `` what are you doing here? '' i asked.  he looked at me. he looked like he was about to say something. `` i need you to come with me. '' he said, and i followed him.  we were in the middle of nowhere when the man shot me in the head. i looked at him, and he looked back at me, then at the gun. i felt a sharp pain in my chest. i started to cry. i tried to speak, but my throat was dry. i heard the man say something, and then he shot me. i fell to the ground, bleeding. i woke up in a hospital bed. there was a note on my bedside table. i opened it and read it aloud.  *i'm sorry, but this isn't what i was expecting. \\

      \rowcolor{light-gray} \multirow{-1}{*}{\textsc{RL}} & what's that?  \hl{we've been waiting for you for a long time, i said.} i'm not sure how long you'll be waiting for me. you don't have much time to think about it, do you? i mean, i can see it, but i can feel it in your eyes, and i can hear it in the way you can hear me. i can smell it, and it can feel your heart beating faster than you can think. i think you know what i am talking about. i know what you are talking about, but you can not understand what i mean. i want you to know that i love you. i need you to understand that you are my love, and that i will always be there for you. \hl{that you will always love me. that i am the love of my life, and you are the love that i have always wanted to be.} and i will never let you down. i will not let you go. i promise you that.with that, i walked out of the building, into the street, and into the night.\\
    \bottomrule
 	\toprule
 	\textbf{Input} &  100000 year high\\     
 	\textbf{prompt} & \temprel{before} \temprel{after} \temprel{before} \temprel{before}  \\
 	\midrule
 	\rowcolor{light-gray} \multirow{-1}{*}{\textbf{baseline}} & for a long time it seemed the world 's people all were living in the same way . the same . they were all the same . the same , they had the same families and their children and they all had the same friends . but these families were all different . they may have been the last of them , and the world was a better place without it , but those people knew about it , they just did n't know the truth . some of them lived for years , some of them a century . and a new age started , one that would make this world a better places . a world without war or war . it seemed the first people could see the world and see the future and people living it . they were the first to do it . '' `` the first people on earth who could live a normal life ! the first to have a normal family and a normal home ! the first ones who we could live on ! who could have imagined that our world was as good as ours was and that the next life would be better than what i ' ve been , that there 's a whole new life ahead of us , that this world is a living one ! '' and that 's when it happened . that was the day . on march 20th , 2036 . march 20, 2036 , the first time that day\\

      \multirow{-1}{*}{\textsc{Vanilla}} & it was a beautiful day. the sun was shining and the sky was blue. the clouds were bright and the trees were green and the birds were singing. it was a wonderful day.  i was walking down the street when i saw it. the sky above me was blue and it was beautiful. i had never seen anything like it before. i looked down at the ground and saw nothing. i didn't know what to do. i took a step back and looked up at the sky. i couldn't see anything at all. i walked towards the sky and looked at the stars. they were beautiful. they seemed like they were floating. i reached out and touched them. i felt a warmth in my chest. i touched them again and again.  it was amazing. i wasn't sure if i was dreaming or not, but i knew i was.\\

      \rowcolor{light-gray} \multirow{-1}{*}{\textsc{RL}} & it's been a long, long time since i've seen the light of the sun. i'm not sure how long it has been, but i think it's a long time. \hl{i 'd like to say that the last time i saw the light, it was a thousand years ago, but it wasn't.}  it was only a few years ago that i first saw the sun, when it was just a few seconds away from reaching the horizon. i don't know how long i have been seeing it, but the light is blinding, and i can't help but wonder what it could have been like.  maybe i 'll see it again, but for now, i just want to see the light again.\\
    \bottomrule

  	\bottomrule
 	\textbf{Input} & paint a picture of loneliness , without using the word , or any of its forms . \\     
 	\textbf{prompt} & \temprel{before} \temprel{vague} \temprel{vague} \temprel{vague} \temprel{before} \temprel{before} \temprel{before} \temprel{after} \temprel{after} \temprel{before} \\
 	\midrule
 	\rowcolor{light-gray} \multirow{-1}{*}{\textbf{baseline}} &  the glass is cold , but the sun shines brightly through . on the glass , in a small room , a woman stands . it stands before her , its hands clutching the glass in front of it like a shield . it 2019 s not the sunlight , but it hits the sky like a storm . the light makes the clouds roll in from all directions . a soft breeze , the wind blowing through the trees , the water rustling in the water . there are people in the room , but most do n't look at their feet , they 're too busy staring out the window . some are silent . most look at the woman . one of the people looks back at the woman who seems to be alone in this room , who is alone with a glass of champagne . the wind and water make the trees sway . you hear the birds chirping from outside . a few laugh , some are sad , some angry , but most are happy , and some are happy . there 's a small child sitting on her lap . in the corner . on the ground , on a bench . and it 's just this woman and the glass .\\

      \multirow{-1}{*}{\textsc{Vanilla}} & i've always been lonely. i'm not sure why. maybe it's the fact that i don't have any friends. maybe i've been alone for a long time.  i haven't seen anyone. i can't even remember where i am. i have no friends. there's no one to talk to. no one who can help me. i just want to be alone.  sometimes i think that i 'll just be alone forever.  but then i think back to the first time i saw someone else. it was a girl. she's beautiful, but she looks like she doesn't belong here. she has no friends, no family, no one.  then i thought of her. i think of the way she looked at me. the way i looked at her, the way her eyes looked at the people around her.  that's when i realized that i was alone.\\

      \rowcolor{light-gray} \multirow{-1}{*}{\textsc{RL}} & i don't know why i'm doing this, but i've been doing it for a long time now.  it's been a long, long time since i 'd seen a single person.  i remember the first time i saw her. she was beautiful, and i loved her.  but i couldn't tell you how much i missed her. i didn't want to know how much she meant to me.  \hl{and i can't even remember the last time i talked to her, because i was too busy trying to figure out what to say.}  \hl{she was always there for me, and she always made me feel so alone.}  sometimes i wish i could go back, but sometimes i think i 'll just stay here forever.  maybe someday, when she's gone, she 'll come back.\\
 	
 	\bottomrule
 	\end{tabular}
 	}
   	\caption{Additional examples for WritingPrompt. Possible \temprel{after} relations are highlighted.}
   	\label{tab:wp-examples-more}
 \end{table*}

\clearpage
\begin{figure*}[t!]
    \centering
\includegraphics[width=\textwidth]{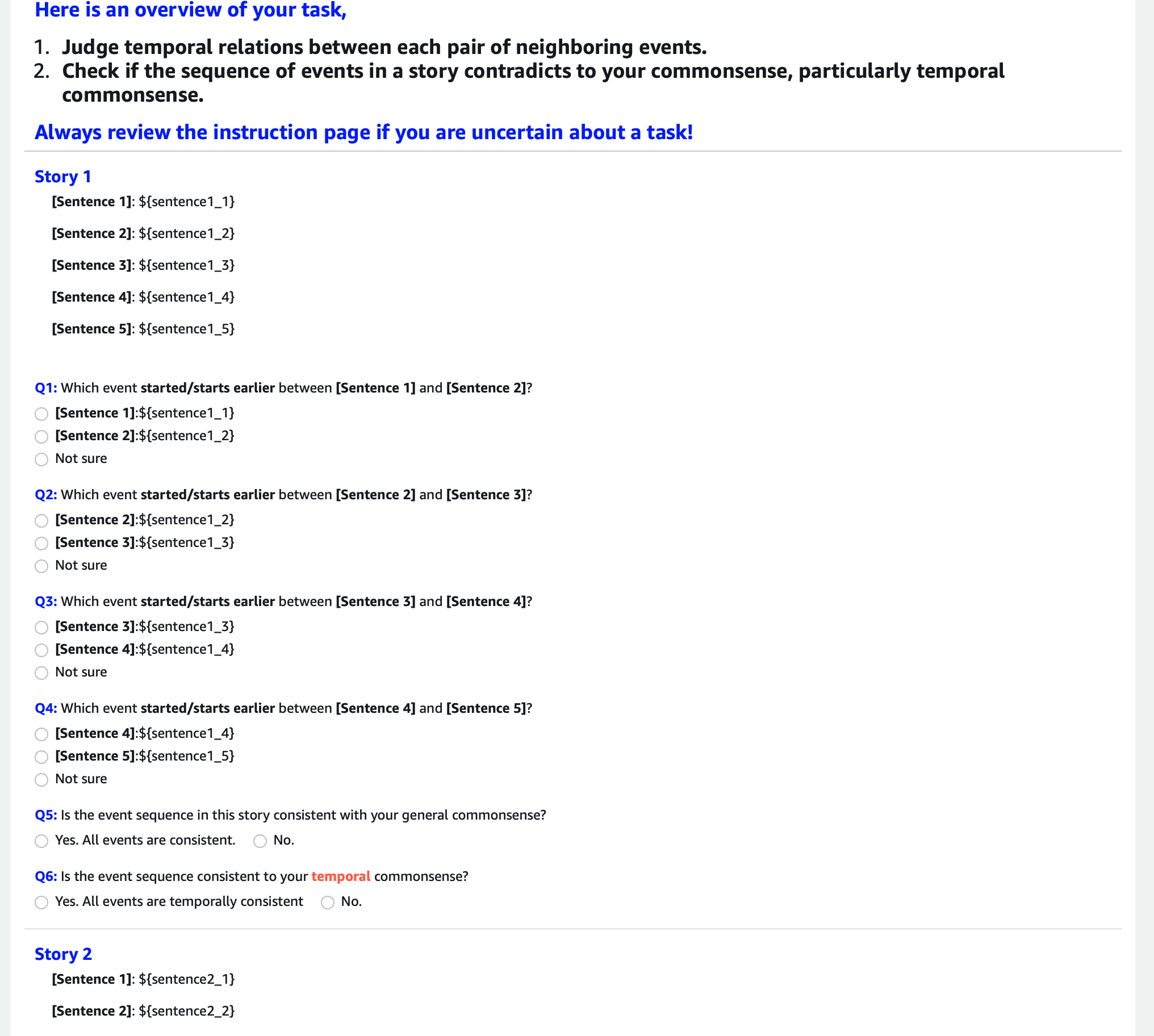}
\caption{Annotation interface for pairwise event temporal relations and temporality.}
\label{fig:annotation_1}
\end{figure*}

\clearpage

\begin{figure*}[t!]
    \centering
\includegraphics[width=\textwidth]{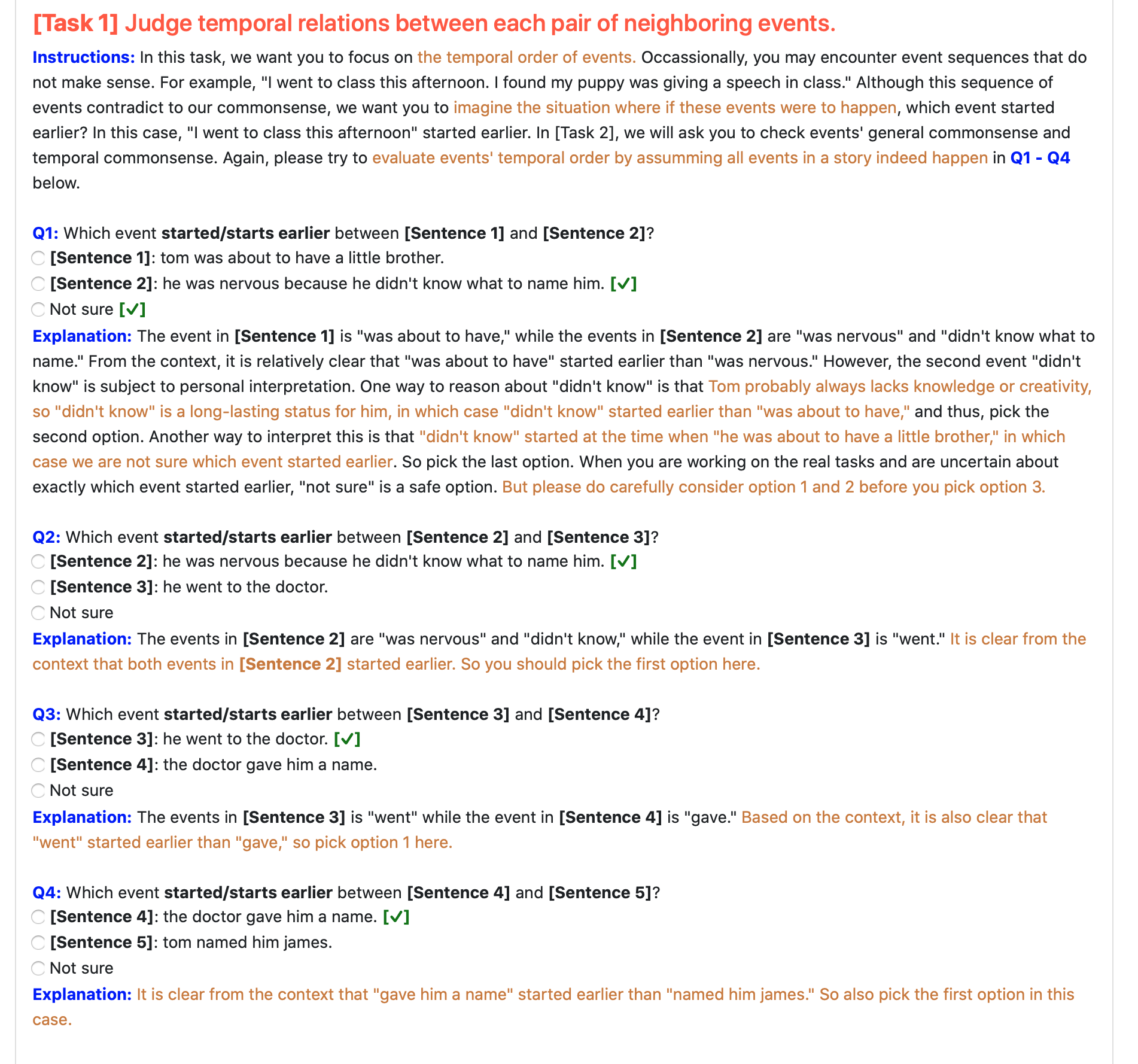}
\caption{Instruction for pairwise event temporal relation annotation.}
\label{fig:instruct_1}
\end{figure*}

\clearpage

\begin{figure*}[t!]
    \centering
\includegraphics[width=\textwidth]{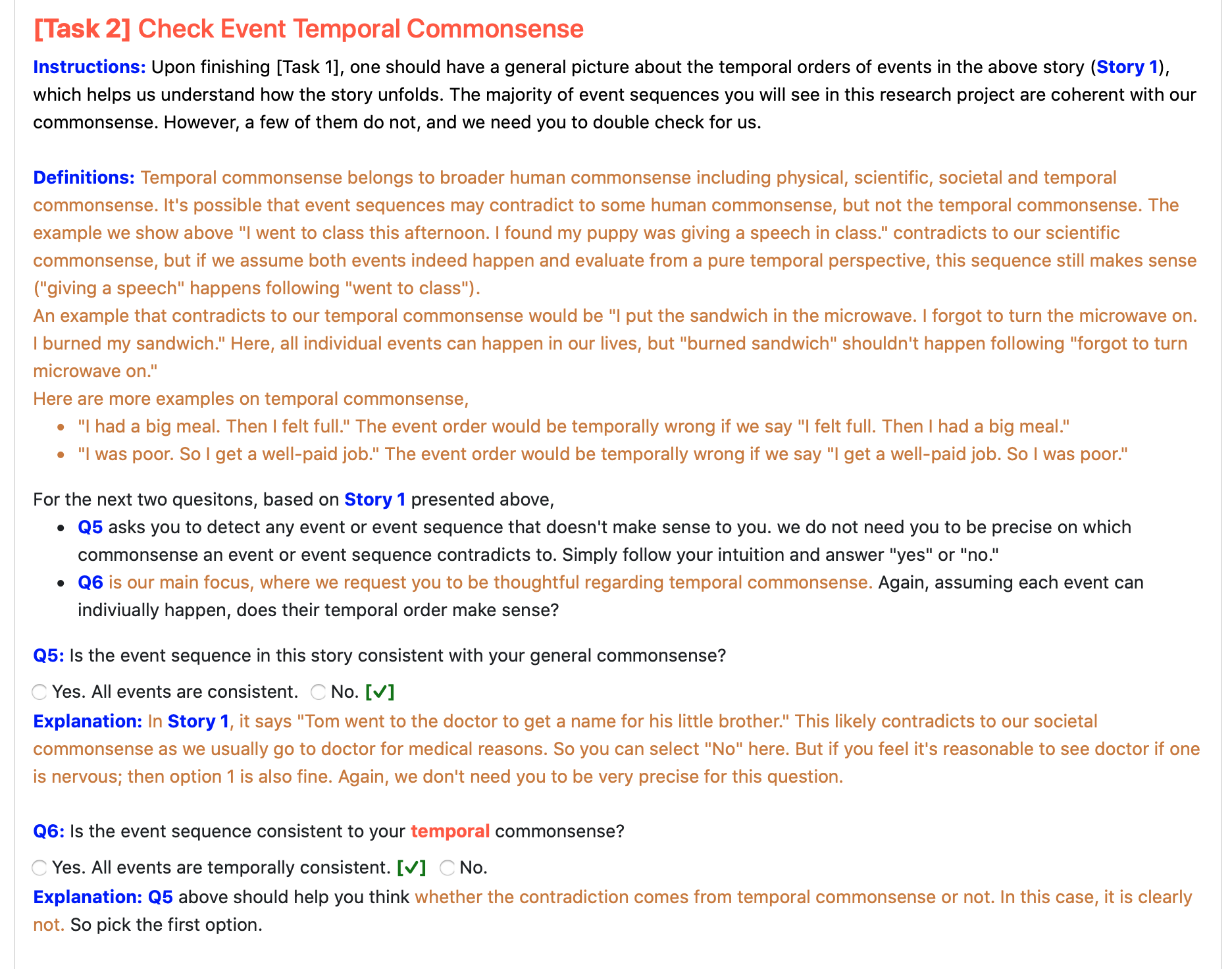}
\caption{Instruction for temporality annotation.}
\label{fig:instruct_2}
\end{figure*}

\clearpage

\begin{figure*}[t!]
    \centering
\includegraphics[width=\textwidth]{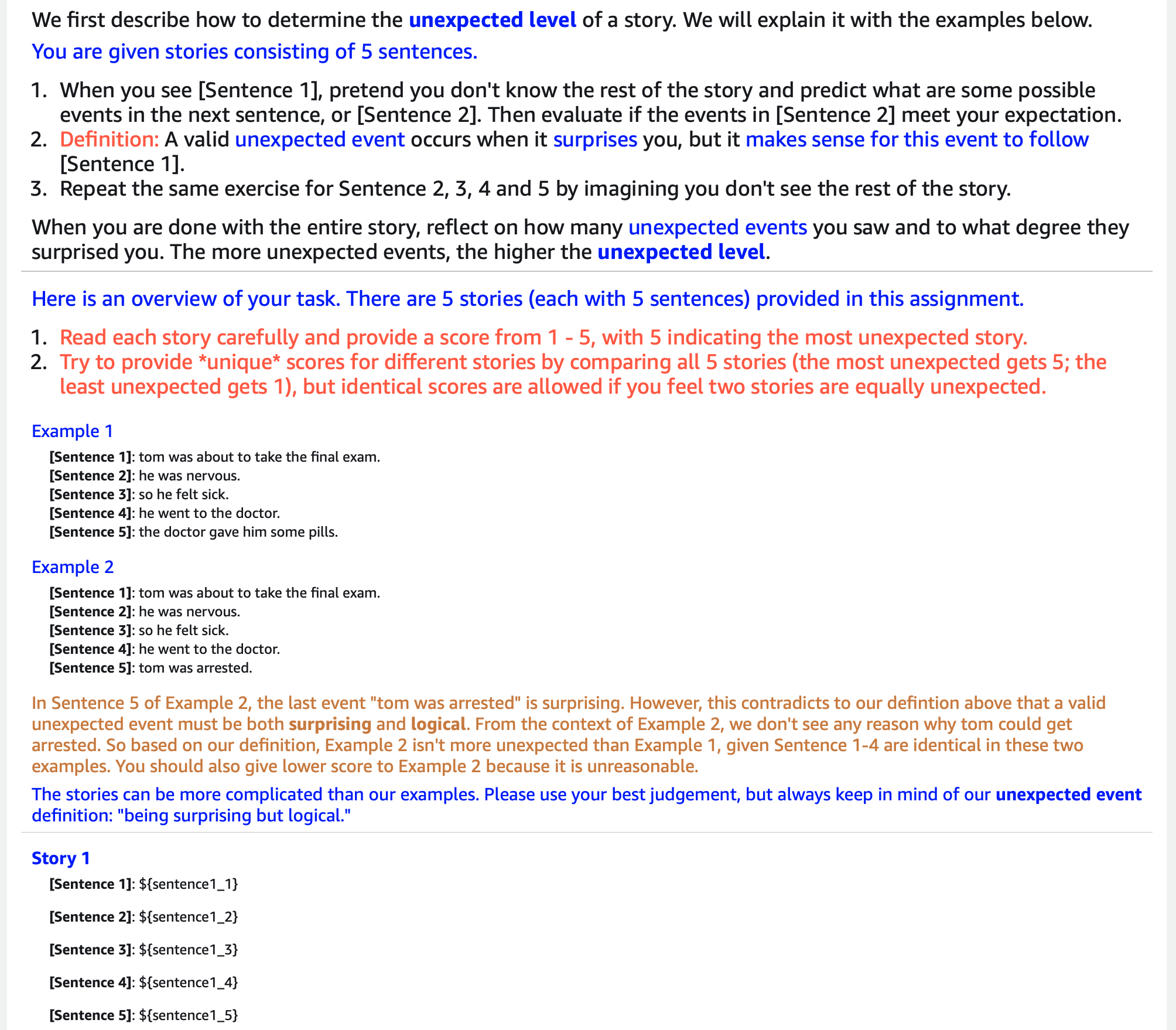}
\caption{Annotation interface for interest level (unexpectedness).}
\label{fig:annotation_2}
\end{figure*}

\end{document}